%% file: main.tex
\documentclass{article} %
\usepackage{iclr2027_conference,times}

\input{imports.tex}

\input{definitions.tex}

\input{math_commands.tex}

\usepackage[colorlinks,citecolor=blue!60!black,linkcolor=red!60!black,urlcolor=magenta!60!black]{hyperref}
\usepackage{url}
\usepackage{marvosym}
\usepackage[capitalize,nameinlink]{cleveref}

\makeatletter
\newcommand{\appendixtoc}{\begingroup\small\parskip 0pt\hypersetup{linkcolor=black}\@starttoc{toc}\endgroup}
\makeatother

\usepackage{etoolbox}
\AtBeginEnvironment{table}{\setlength{\abovecaptionskip}{0pt}\setlength{\belowcaptionskip}{4pt}}
\makeatletter
\g@addto@macro\normalsize{\abovedisplayskip 4pt plus 2pt minus 3pt \belowdisplayskip 4pt plus 2pt minus 3pt}

\makeatother

\title{\textbf{COSMI}: \textbf{\underline{CO}}mpositional \textbf{\underline{S}}ynthesis of \textbf{\underline{M}}ulti-Object \textbf{\underline{I}}nteractions}

\author{
\begin{tabular}{@{}l@{\hspace{1.4em}}l@{\hspace{1.4em}}l@{}}
Daniel Eskandar\textsuperscript{1,2,3\,*} & Ilya A. Petrov\textsuperscript{1,2\,*\,\Letter} & Gerard Pons-Moll\textsuperscript{1,2,4}
\end{tabular}
\vspace{-4mm}
\\ \\
{\small \textsuperscript{1}University of Tübingen, Germany \quad \textsuperscript{2}Tübingen AI Center, Germany \quad \textsuperscript{3}Zuse School ELIZA, Germany}\\
{\small \textsuperscript{4}Max Planck Institute for Informatics, Saarland Informatics Campus, Germany}\\
{\small \url{https://ptrvilya.github.io/cosmi/}}
}

\iclrfinalcopy %
\begin{document}
\addtocontents{toc}{\protect\setcounter{tocdepth}{-1}}
\maketitle
\lhead{Preprint}
{\renewcommand{\thefootnote}{*}\footnotetext{Equal contribution.}\renewcommand{\thefootnote}{\Letter}\footnotetext{Corresponding author: \href{mailto:ptrvilya@gmail.com}{ptrvilya@gmail.com}.}}

\begin{figure}[H]
    \centering
    \vspace{-15pt}
    \includegraphics[width=1.0\linewidth]{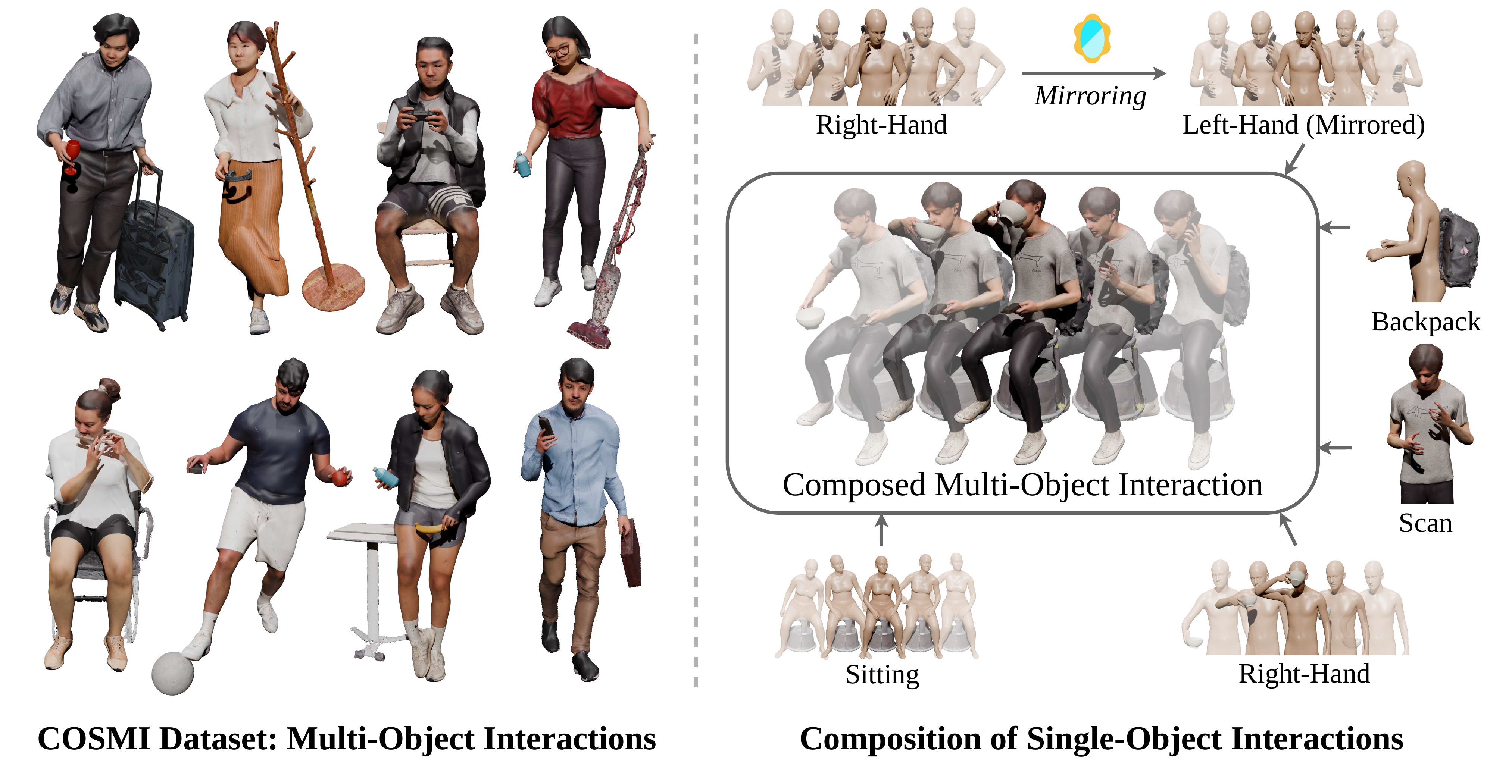}
    \vspace{-20pt}
    \caption{\textbf{\methodname{}.} Left: sequences of the \datasetname{}, composed from single-object captures and rendered with clothed scans. Right: composition of one sequence from a sitting clip, a right-hand action, a left-hand action obtained by mirroring, and a backpack, rendered with a clothed scan.}
    \label{fig:teaser}
\end{figure}

\begin{abstract}
Generative models of human-object interaction are bounded by the data that exists: everyday activities involve several objects, but most captured datasets record one at a time, as multi-object capture is combinatorially expensive.
Our observation is that interactions are local, so single-object captures already contain the parts of multi-object activities.
We compose them: contact-consistent clips of single interactions, mirrored to balance the hands, transfer between bodies, and a language model and geometric checks admit only the pairings that are plausible, semantically and physically.
Therefore, the dataset grows combinatorially with the clips rather than recording time.
The \datasetname{} holds 222k sequences and 275 hours with up to five objects, nearly thirty times the largest multi-object capture, and can be extended by adding datasets or even hand-object recordings.
On this data we train the \methodname{} method, a text-to-interaction diffusion transformer that follows how the data is built: weight-shared object slots generate a variable number of objects, predicted relative to the body parts that move them.
On a benchmark with an unseen object and unseen interaction combinations, models trained on the dataset generalize to the unseen combinations.
\methodname{} outperforms baselines in text alignment and contact accuracy, where its margin is largest on the unseen object.
Code, models, and the dataset pipeline will be released.
\end{abstract}

\section{Introduction}
\label{sec:intro}
\input{sections/01_introduction}
\section{Related Work}
\label{sec:related}
\input{sections/02_related_work}
\section{The \methodname{} Dataset}
\label{sec:dataset}
\input{sections/03_dataset}
\section{The \methodname{} Method}
\label{sec:method}
\input{sections/04_method}
\section{Experiments}
\label{sec:experiments}
\input{sections/05_experiments}
\section{Conclusions}
\label{sec:conclusions}
\input{sections/06_conclusions}
\input{sections/07_statements_arxiv}

\bibliography{main}
\bibliographystyle{iclr2027_conference}

\clearpage
\appendix
\addtocontents{toc}{\protect\setcounter{tocdepth}{3}}
\setcounter{topnumber}{3}
\setcounter{totalnumber}{4}
\input{sections/08_appendix}

\end{document}

%% file: imports.tex
\usepackage{graphicx}
\usepackage{booktabs}

\usepackage[accsupp]{axessibility}  %

\usepackage{xcolor}
\usepackage{multicol,multirow}
\usepackage{overpic}
\usepackage{float}
\usepackage{wrapfig}
\usepackage{amssymb}
\usepackage{afterpage}
\usepackage{pifont}%
\usepackage{wrapfig}
\usepackage{makecell}
\usepackage{dsfont}
\usepackage{nccmath}
\usepackage{mathtools}
\usepackage{nicefrac}
\usepackage{listings}
\usepackage{rotating}
\usepackage{enumitem}
\usepackage{adjustbox}
\usepackage{tcolorbox}
\tcbuselibrary{skins,breakable}

%% file: definitions.tex
\definecolor{amber}{HTML}{CC7A00}
\definecolor{tok_col}{RGB}{92,127,146}    %
\definecolor{word_col}{RGB}{217,160,91}   %
\definecolor{t_col}{RGB}{150,120,130}     %
\newcommand{\coltok}[1]{\textbf{\textcolor{tok_col}{#1}}}
\newcommand{\colword}[1]{\textbf{\textcolor{word_col}{#1}}}
\newcommand{\colt}[1]{\textbf{\textcolor{t_col}{#1}}}
\newcommand{\cmark}{\textcolor{green!60!black}{\ding{51}}}
\newcommand{\xmark}{\textcolor{red!70!black}{\ding{55}}}
\newcommand{\na}{\textcolor{gray}{--}}
\newcommand{\dgood}[1]{\,{\scriptsize\textcolor{green!50!black}{$(-#1)$}}}
\newcommand{\dmild}[1]{\,{\scriptsize\textcolor{amber}{$(+#1)$}}}
\newcommand{\dsev}[1]{\,{\scriptsize\textcolor{red!70!black}{$(+#1)$}}}
\newcommand{\dgray}[1]{\,{\scriptsize\textcolor{black!55}{$(+#1)$}}}
\newlength{\pvw}
\newlength{\pdw}
\newcommand{\cw}[2]{\settowidth{\pvw}{0.00}\settowidth{\pdw}{\dmild{0.00}}\makebox[\pvw][r]{#1}\makebox[\pdw][l]{#2}}

\newcommand{\methodname}{COSMI}
\newcommand{\datasetname}{\methodname{} dataset}

\newcommand{\mat}[1]{\boldsymbol{#1}}
\renewcommand{\vec}[1]{\boldsymbol{#1}}
\newcommand{\set}[1]{\mathcal{#1}}
\newcommand{\real}{\mathbb{R}}
\newcommand{\obj}{\set{O}}
\newcommand{\bodyj}{\set{J}}
\newcommand{\mir}{\mat{M}}

\newcommand{\first}[1]{\textbf{#1}}
\newcommand{\second}[1]{\underline{#1}}
\newcommand{\err}[1]{$^{\scriptscriptstyle\pm#1}$}
\newcommand{\noerr}[1]{\phantom{\err{#1}}}
\newlength{\meanwidth}
\newcommand{\nomean}[1]{\settowidth{\meanwidth}{#1}\makebox[\meanwidth]{--}}
\newcommand{\overmean}[2]{\settowidth{\meanwidth}{#2}\makebox[\meanwidth]{#1}} %

\newcommand{\nobj}{N}
\newcommand{\window}{W}
\newcommand{\methodnetwork}{\mathrm{COSMI}_\psi}
\newcommand{\cond}{\set{C}}
\newcommand{\textok}{\set{W}} %

\newcommand{\h}{\set{H}}

\newcommand{\hpose}{\vec{\theta}_{\h}}
\newcommand{\hhands}{\vec{\theta}_{\mathrm{hands}}}
\newcommand{\hjoints}{\mat{J}_{\h}}
\newcommand{\hcontact}{\vec{c}_{\h}}

\newcommand{\objglobal}{\mat{T}_{\obj}}
\newcommand{\objrel}[1]{\mat{T}_{\obj|#1}}
\newcommand{\objhinge}{\vec{a}_{\obj}}
\newcommand{\objcontact}{\vec{c}_{\obj}}
\newcommand{\relj}{\bodyj_{\mathrm{rel}}}
\newcommand{\loss}[1]{\mathcal{L}_{\mathrm{#1}}}

%% file: math_commands.tex
\usepackage{amsmath,amsfonts,bm}

\def\eqref#1{equation~\ref{#1}}

\def\1{\bm{1}}

\DeclareMathAlphabet{\mathsfit}{\encodingdefault}{\sfdefault}{m}{sl}
\SetMathAlphabet{\mathsfit}{bold}{\encodingdefault}{\sfdefault}{bx}{n}

%% file: sections/01_introduction.tex
Generating 3D human-object interaction (HOI) serves animation, embodied AI, and augmented reality, from humanoid controllers and robot policies that learn from HOI motion~\citep{xu2025intermimic,wang2025skillmimic} to egocentric assistants~\citep{banerjee2025hot3d,ma2024nymeria,petrov2026echo} and composite animation such as sitting down while carrying an object~\citep{pan2025tokenhsi}.
The activities these applications target are frequently multi-object: in the KIT whole-body database, 99 of the 262 object-related motions involve two or more objects~\citep{mandery2016unifying}.
Like other generative models, motion generators improve with data~\citep{wang2025scaling,xu2025interact}, so what HOI generators can produce is bounded by the interaction data that exists.

On the data side, HOI datasets mostly record one person with one object~\citep{taheri2020grab}, and the generators trained on them inherit this scope~\citep{xu2023interdiff,diller2024cghoi,peng2025hoidiff}.
Capturing several objects at once is expensive, and the cost grows combinatorially with the objects, body parts, and pairings~\citep{xie2024procigen}, so captured multi-object datasets stay below ten hours~\citep{lv2024himo,kim2025parahome,lu2025humoto}, against 21.8 in the single-object collection InterAct~\citep{xu2025interact}.
On the model side, generators are built for one object~\citep{diller2024cghoi,pi2025coda,xu2023interdiff}, and the few multi-object ones train one model per object count~\citep{lv2024himo,he2025syncdiff} or pad a fixed state~\citep{zhang2024hoim3} and leave implicit which body part moves which object.
The concurrent SK-HOI~\citep{peng2026surface} handles a variable number of objects in three separate stages.
There is no multi-object data at single-object scale, and no end-to-end generator for a variable number of objects grounded to the body parts that move them.

Synthetic and composed data has closed comparable gaps before~\citep{black2023bedlam,xie2024procigen,mandlekar2023mimicgen}, and closest to our setting, SINC composes simultaneous actions by taking each body part from the clip whose action uses it~\citep{athanasiou2023sinc}.
Our observation is that interactions with objects are local in the same way: a cup is held by a hand, a chair supports the pelvis, a ball is kicked by a foot, and the rest of the body is free.
The limb that performs one interaction can therefore be transferred, together with its object, to a body that performs another, and multi-object interactions can be composed from single-object captures (\cref{fig:teaser}), which exist at scale.
The composition is only as good as its gating: from contact-consistent clips of single interactions, a language model decides which can plausibly happen at once, and geometric checks keep only the compositions that stay in contact and free of penetration (\cref{fig:composition}).
Mirroring the one-handed and one-footed clips balances the left and right hands, which captured interactions skew to the right.
The resulting \datasetname{} contains 222k sequences and 275 hours of motion, nearly thirty times the largest captured multi-object dataset~\citep{lv2024himo}.
Its physical plausibility stays close to the parent captures (\cref{sec:app_quality}), and since nothing in the composition is specific to a dataset, it grows by adding a rule rather than a recording session, even from hand-object recordings (\cref{sec:experiments}).

Such data calls for a generator for a variable number of objects.
\methodname{} is a single diffusion transformer that denoises the person and a set of objects from a prompt and initial poses.
Its object slots share their weights, so no parameter depends on the object count, and one model generates one to five objects.
Each object is predicted relative to the joint that drives it, which a part-assignment head reads from these conditions, and text tokens link the words of the prompt to the objects.
On a benchmark that holds out an unseen object and five unseen combinations, such as walking-with-an-apple (while walking-only and apple-eating stay in training), \methodname{} and the baselines trained on the \datasetname{} alike generalize to the unseen combinations.
\methodname{} outperforms HIMO~\citep{lv2024himo} and multi-object adaptations of MDM and PriorMDM~\citep{tevet2023mdm,shafir2024priormdm} in text alignment and contact accuracy, where its margin is largest on the unseen object.

In summary, our contributions are:
\begin{itemize}[leftmargin=*,itemsep=1pt,topsep=2pt]
    \item Composition of multi-object interactions from single-object captures, and the \datasetname{}: 222k sequences with up to five objects, left and right balanced by mirroring, extensible by design.
    \item The \methodname{} method: a text-conditioned diffusion transformer for a variable number of objects, with joint-relative prediction, part assignment, auxiliary losses, and contact-consistent guidance.
    \item A benchmark on the \datasetname{} that tests generalization to an unseen object and to unseen interaction combinations. Code, models, and the dataset pipeline will be released.
\end{itemize}

%% file: sections/02_related_work.tex
\paragraph{Human-object interaction datasets.}
Most full-body interaction datasets, captured with marker-based motion capture or multi-view video, record one person with one object~\citep{taheri2020grab,bhatnagar2022behave,huang2024intercap,li2023omomo,zhang2023neuraldome,zhao2024imhoi}, later also with articulated objects~\citep{jiang2023chairs,fan2023arctic}.
Full-body datasets with several objects per sequence are recent and small (\cref{tab:dataset-comparison}): HIMO records two or three small rigid objects~\citep{lv2024himo}, ParaHome and HUMOTO articulated household objects~\citep{kim2025parahome,lu2025humoto}, and HOI-M3, CORE4D, and MMHOI several people~\citep{zhang2024hoim3,liu2025core4d,kogashi2026mmhoi}.
Hand-object datasets omit the body but record more hours and objects, often several at once~\citep{hampali2020honnotate,chao2021dexycb,zhan2024oakink2,liu2024taco,banerjee2025hot3d,fu2025gigahands}, and a body fitted to such a recording turns it into a source clip for our pipeline (\cref{sec:experiments}).
Interaction data has also been generated procedurally, placed into scenes, augmented, or recovered from video~\citep{xie2024procigen,wang2022humanise,xu2025interact,zhang2026interpose,xie2026grail}, always with one object.
The \datasetname{} instead composes these single-object captures into multi-object sequences: every source clip pairs with every compatible target and composed sequences serve as targets again, so the number of sequences grows combinatorially with the clips rather than linearly with recording time (\cref{tab:dataset-comparison}).

\paragraph{Compositionality in motion and interaction modeling.}
Animation has long composed captured clips without objects, in time with motion graphs~\citep{kovar2002motion} and across body parts by splicing~\citep{ikemoto2004enriching,heck2006splicing}.
SINC revisits spatial composition with language, composing the body parts of two clips into training data from GPT-3 part labels, and leaves the semantic compatibility to future work~\citep{athanasiou2023sinc}.
Later models annotate, coordinate, or blend parts~\citep{li2026frankenmotion,zou2024parco,jiang2025dynamic} or compose actions in time at inference~\citep{athanasiou2022teach,shafir2024priormdm,petrovich2024stmc,barquero2024flowmdm}.
Language models also plan compositions at inference, as contact chains, joint-contact pairs, affordance graphs, or plans~\citep{xiao2024unified,wang2024intercontrol,li2026hoipage,xu2024interdreamer,sun2026coma}, while we use one offline to infer the semantic plausibility of clip combinations (\cref{sec:dataset}).
Composed training data exists for multi-person motion, by random placement~\citep{shan2024multiperson} or with LLM guidance and a learned filter~\citep{wu2026text2interact}, and for interactions only as static poses~\citep{zhao2022coins} or as segments and models composed at inference~\citep{lv2024himo,na2025hhoi}.
We extend body-part composition to interactions with objects, at the data level and across datasets, with object transfer across bodies, semantic gating, and physical checks.

\paragraph{Text-driven interaction synthesis.}
Text-to-motion generation rests on diffusion and token models trained on HumanML3D~\citep{guo2022humanml3d,tevet2023mdm,zhang2024motiondiffuse,chen2023mld,zhang2023t2mgpt,guo2024momask} and, recently, on million-clip corpora~\citep{wang2025scaling}, two orders of magnitude larger than any captured interaction dataset.
Interaction generators extend them to a body and one object~\citep{ghosh2023imos,xu2023interdiff,li2023omomo,li2024chois,peng2025hoidiff,diller2024cghoi,xu2024interdreamer,petrov2025tridi,ron2026hoidini,ji2026hoimask,pan2026fast,jung2026dechoi,cai2026vihoi}, including articulated ones~\citep{pi2025coda}, and grasp synthesis models the fingers in contact~\citep{taheri2022goal,wu2022saga,zhang2021manipnet,cha2024text2hoi,christen2024diffh2o,zhang2025bimart}.
Generators for several objects are few: HIMO trains separate models for two and three objects~\citep{lv2024himo}, SyncDiff one model per object configuration, conditioned on label phrases~\citep{he2025syncdiff}, and Dex2HOI one stream per object for at most two, both in the hands~\citep{pratikaki2026dex2hoi}.
HOI-M3 pads a preset state and takes no text~\citep{zhang2024hoim3}, and HOI-PAGE optimizes each sequence zero-shot for minutes~\citep{li2026hoipage}.
The concurrent SK-HOI covers a variable number of objects with zero-padded slots, but generates in three separately trained stages followed by minutes of test-time optimization per sequence~\citep{peng2026surface}.
\methodname{} instead is one text-conditioned diffusion transformer with no parameter tied to the object count, which generates one to five objects, articulated and worn, in a single stage and grounds the objects to the body parts that move them.

%% file: sections/03_dataset.tex
\begin{figure}[t]
    \centering
    \vspace{-5pt}
    \includegraphics[width=1.0\linewidth]{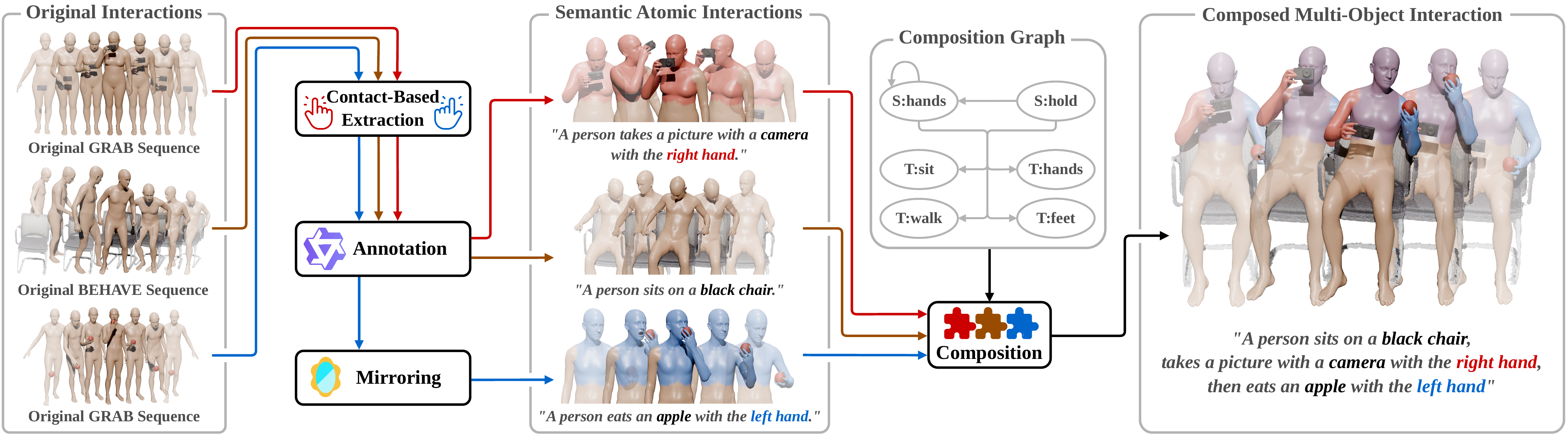}
    \vspace{-10pt}
    \caption{\textbf{Composition of one sequence.} Atomic clips are extracted by interaction rules, annotated by a language model, and mirrored. A sitting target receives two compatible hand actions, their limbs and objects are transferred onto its body, and the caption is assembled accordingly.}
    \vspace{2pt}
    \label{fig:composition}
\end{figure}

We build every sequence of the \datasetname{} from captured single-object interactions and motions, combined on one body.
Our goal is to obtain semantically plausible multi-object interactions from these atomic clips.
We therefore extract contact-consistent clips from the source sequences, annotate them with semantic information from a language model, and mirror them to balance the left and right hands.
To combine the clips, we rely on this semantic information to pair plausible matches, followed by geometric checks of the physical plausibility of the result (\cref{fig:composition}).
We build on GRAB~\citep{taheri2020grab}, BEHAVE~\citep{bhatnagar2022behave}, InterCap~\citep{huang2024intercap}, OMOMO~\citep{li2023omomo}, and ARCTIC~\citep{fan2023arctic}, with BEHAVE and InterCap in their InterAct~\citep{xu2025interact} versions, and on CMU walking sequences from AMASS~\citep{mahmood2019amass} selected with BABEL labels~\citep{punnakkal2021babel}.
All sources are unified to a gender-neutral SMPL-X~\citep{pavlakos2019smplx} body at 30\,fps in a common frame, and every object is a template mesh with a rigid transform per frame, one per part for articulated objects (\cref{sec:app_preprocessing}).

\paragraph{Contact-based extraction.}
\label{sec:dataset_extraction}
We build the dataset from \emph{atomic clips}, segments in which a person performs a single interaction and stays in contact with the object from the first frame to the last, so that the acting limb and the object can be lifted out with their contact intact.
Thirteen \emph{interaction rules} specify the interaction types by the body parts that must and must not touch the object, by pose constraints such as an upright torso, and by the \emph{local reference} joints relative to which the object is expressed (\cref{sec:app_rules}).
We evaluate every rule per frame, with contact defined as proximity below 5\,cm to the object surface, and take contiguous windows of at least 3\,s as atomic clips, 1,491 in total.
Holds and worn objects reduce to a single representative frame, and the walking rule admits whole sequences.
In composition, \emph{source} rules cover the manipulation of small objects with one or both hands and passive holds (GRAB, ARCTIC).
\emph{Target} rules cover the interactions that engage the whole body, sitting, moving large objects, foot interactions, and walking (BEHAVE, InterCap, OMOMO, AMASS), and \emph{augmentation} rules the worn backpack and eyeglasses (BEHAVE, GRAB).

Global trajectories do not transfer between bodies, so we express each object relative to the body part performing its interaction, the local reference joints $\bodyj$ its rule specifies, such as the wrist or the foot.
For a joint $j \in \bodyj$ with world rotation $\mat{R}^t_j$ and position $\vec{t}^t_j$, the relative transform is
\begin{equation}
    \mat{R}^t_{\obj|j} = (\mat{R}^t_j)^{\top} \mat{R}^t_\obj, \qquad
    \vec{t}^t_{\obj|j} = (\mat{R}^t_j)^{\top} (\vec{t}^t_\obj - \vec{t}^t_j),
    \label{eq:relative}
\end{equation}
and inverting it with the joints of \emph{any} body carries the object over to a mirrored or a composed body.

\paragraph{Annotation.}
\label{sec:dataset_annotation}
Whether two interactions can plausibly be performed simultaneously is a semantic question: a person can browse a phone while sitting, but not browse a phone while taking a picture with a camera, although one hand is free for each.
Composition therefore needs to know what every clip captures, which the labels of the source datasets, written for whole recordings in differing forms, do not tell.
We extract this information with a language model, Qwen3~\citep{yang2025qwen3}, which names the sustained interaction with a verb, correcting labels that are wrong or too general, and writes one sentence, such as \emph{``A person drinks from a cup with the right hand.''} (prompts in \cref{sec:app_prompts}).
Per action and object, it further marks whether the action \emph{requires gaze} at the object and \emph{involves the head}, as drinking, eating, and calling do, which decides what can happen at once.

\paragraph{Mirroring.}
\label{sec:dataset_mirroring}
Captured interactions favor the right hand (\cref{tab:dataset-comparison}), so we mirror every one-handed and one-footed clip and swap its rule, which equalizes the two sides and brings the number of clips to 2,663.
Reflecting the body is straightforward, $\tilde{\mat{R}}^t_j = \mir \mat{R}^t_j \mir$ with $\mir = \mathrm{diag}(-1,1,1)$, followed by swapping the left and right joints.
The object cannot be reflected in the same way: a reflected mesh is a new object, and reflecting only its world pose breaks the contact with the mirrored hand, because the SMPL-X body is not symmetric.
We keep the template $\mat{V}_\obj$ and fit a proper rotation $\mat{Q}_\obj$ and translation $\vec{p}_\obj$ that pose it like its reflection,
\begin{equation}
    \mat{Q}_\obj (\mat{V}_\obj - \bar{\vec{v}}_\obj) + \vec{p}_\obj \;\approx\; \mir (\mat{V}_\obj - \bar{\vec{v}}_\obj),
    \label{eq:mirror_align}
\end{equation}
with $\bar{\vec{v}}_\obj$ the centroid of the template, which is exact for symmetric objects.
Composed with the reflected joint-relative pose, this alignment yields a proper transform of the original mesh that preserves the contact, and \cref{eq:relative} places the object on the opposite limb (details in \cref{sec:app_mirroring}).

\input{tables/01_dataset_comparison}

\paragraph{Composition.}
\label{sec:dataset_composition}
We pair a \emph{target} clip, which provides the body motion, with a \emph{source} clip, whose action is added on a free limb, if the two are compatible at two levels.
The composition graph settles which interactions can share a body (\cref{fig:composition}): whole-body targets accept every source, and a one-hand target or a hand action accepts the other hand.
Within a permitted pair, the semantic annotations decide whether the two actions can plausibly happen at once: a person cannot attend to two objects simultaneously, so two actions that both require gaze or both involve the head follow each other in sequence, a right- and a left-hand action on a sitting, walking, or standing body with a blended transition.
Other pairs are composed simultaneously: the source limb, with the head and spine when the action needs them, replaces its counterpart on the target body, and its object follows through \cref{eq:relative}.
We keep a composition only if it is physically plausible: every object keeps the contacts of its rule, the body does not penetrate itself, and no object penetrates another (\cref{sec:app_merge}).
Enumerating the graph exhaustively, with composed sequences serving as targets again, yields one- to three-object levels and a sequential level, and worn objects are added at every level, up to five objects per sequence.
Captions are assembled from the annotations by a template and paraphrased.

\paragraph{Statistics and quality.}
The \datasetname{} contains 221,677 sequences and 275 hours of motion from 59 subjects and 36 objects, seven of them articulated, with up to five objects per sequence and left and right hands balanced at 49/51 (\cref{tab:dataset-comparison}).
Every composed sequence inherits the capture quality of its parents, which we verify with the metrics of HUMOTO~\citep{lu2025humoto} on the composed sequences and on their parent recordings (\cref{sec:app_quality}).
Foot sliding is unchanged, since composition alters neither the lower body nor the global trajectory, and jerk and penetration change by less than the captured datasets differ among themselves, human jerk the most, by up to half.

\paragraph{Extensibility and benchmark.}
Nothing in the composition is specific to a source dataset: the rules are stated by contacts, pose constraints, and reference joints on the unified body, so any clip on this body enters through a rule, as the AMASS walking sequences do.
Partial observations enter as well, since every interaction is stored relative to the body part that performs it.
A hand-object recording becomes a source clip once a body is fitted to it, which is less obvious, so we demonstrate it with HOT3D (\cref{sec:experiments}).
The exhaustive enumeration also makes generalization testable: a pairing such as walking with an apple can be held out as a whole while walking and eating an apple remain in training.
Our benchmark builds on this and tests how well a model generalizes to five such unseen combinations and to an unseen object, held out with all its sequences (\cref{sec:experiments}).

%% file: tables/01_dataset_comparison.tex
\begin{table}[t]
    \centering
    \caption{\textbf{Dataset comparison.} L\,/\,R: share of left- and right-hand interactions. The \datasetname{} combines the scale of single-object captures with several objects per sequence and balanced hands.}
    \label{tab:dataset-comparison}
    \footnotesize
    \setlength{\tabcolsep}{2pt}
    \renewcommand{\arraystretch}{1.2}
    \begin{adjustbox}{max width=\linewidth}
\begin{tabular}{l c@{\hspace{5pt}} cccc c ccccc}
    \toprule
    \multirow{2}{*}{\textbf{Dataset}}
                        & \multirow{2}{*}[-0.8ex]{%
                        \makecell{\textbf{Multi-}\\\textbf{Object}}}
                        & \multicolumn{4}{c}{\textbf{Interaction Coverage}}
                        & \multicolumn{1}{c}{\textbf{Hand Bias}}
                        & \multicolumn{5}{c}{\textbf{Scale}} \\
      \cmidrule(lr){3-6} \cmidrule(lr){7-7} \cmidrule(lr){8-12}
                        & & \makecell{\textbf{Small-Object}\\\textbf{Manipulation}} & \makecell{\textbf{Large-Object}\\\textbf{Manipulation}} & \makecell{\textbf{Sitting /}\\\textbf{Body Support}} & \makecell{\textbf{Articulated}\\\textbf{Objects}} & \textbf{L\,/\,R (\%)} & \textbf{\# Subjects} & \textbf{\# Objects} & \textbf{\# Sequences} & \textbf{\# Frames (M)} & \textbf{Hours} \\
      \midrule
      GRAB~\citep{taheri2020grab}         & \xmark & \cmark & \xmark & \xmark & \xmark & 15\,/\,85 & 10 & 51 & 1,335   & 1.62  & 3.76   \\
      BEHAVE~\citep{bhatnagar2022behave}  & \xmark & \xmark & \cmark & \cmark & \xmark & 45\,/\,55 & 8  & 18 & 299     & 0.45  & 4.13   \\
      InterCap~\citep{huang2024intercap}  & \xmark & \cmark & \cmark & \cmark & \xmark & 7\,/\,93  & 10 & 10 & 223     & 0.07  & 0.62   \\
      OMOMO~\citep{li2023omomo}           & \xmark & \xmark & \cmark & \xmark & \cmark & 47\,/\,53 & 16 & 15 & 5,280   & 0.96  & 8.91   \\
      ARCTIC~\citep{fan2023arctic}        & \xmark & \cmark & \xmark & \xmark & \cmark & 36\,/\,64 & 9  & 11 & 301     & 0.22  & 2.02   \\
      HIMO~\citep{lv2024himo}             & \cmark & \cmark & \xmark & \xmark & \xmark & 47\,/\,53 & 34 & 53 & 3,376   & 4.08  & 9.44   \\
      HOI-M$^3$~\citep{zhang2024hoim3}    & \cmark & \cmark & \cmark & \cmark & \xmark & 45\,/\,55 & 31 & 67 & 86 & 1.87  & 8.64   \\
      ParaHome~\citep{kim2025parahome}    & \cmark & \cmark & \cmark & \cmark & \cmark & 42\,/\,58 & 38 & 22 & 207 & 0.88  & 8.11   \\
      HUMOTO~\citep{lu2025humoto}         & \cmark & \cmark & \cmark & \cmark & \cmark & 44\,/\,56 & 1  & 63 & 735 & 0.24  & 2.18   \\
      \midrule
      \textbf{COSMI (Ours)} & \cmark & \cmark & \cmark & \cmark & \cmark & 49\,/\,51 & 59 & 36 & 221,677 & 29.74 & 275.4 \\
      \bottomrule
\end{tabular}
    \end{adjustbox}
\end{table}

%% file: sections/04_method.tex
\begin{figure}[t]
    \centering
    \vspace{-5pt}
    \includegraphics[width=1.0\linewidth]{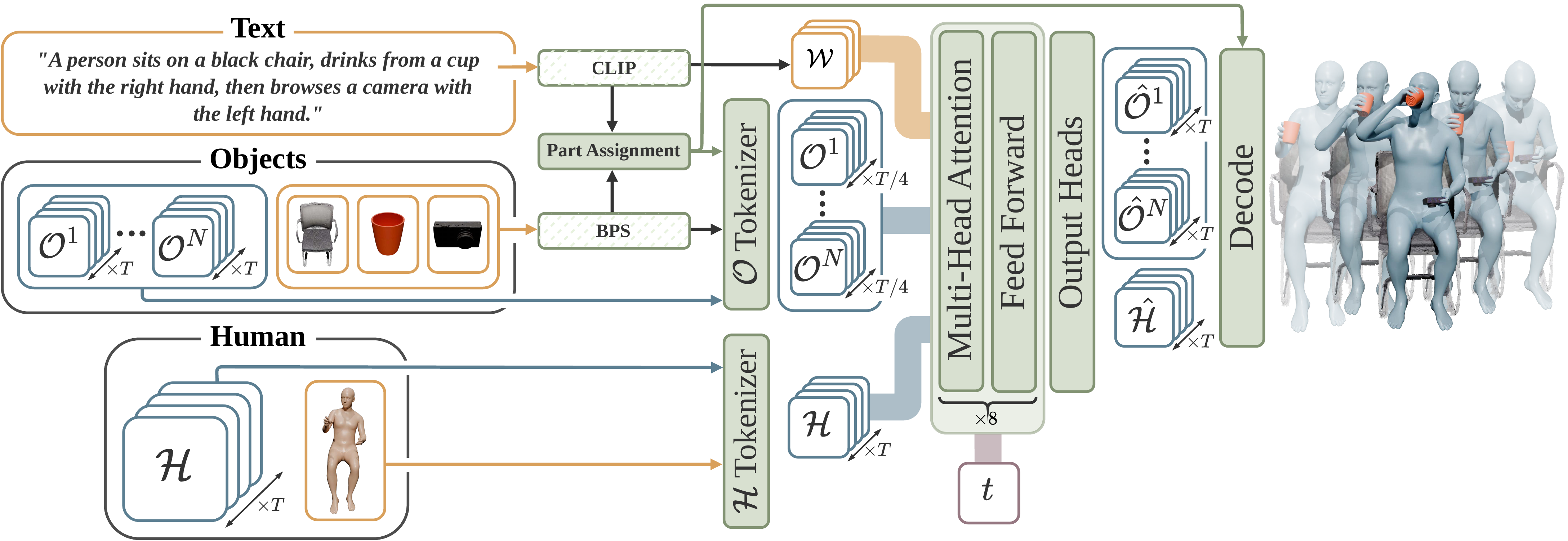}
    \vspace{-10pt}
    \caption{\textbf{\methodname{} method.} One transformer denoises the \coltok{human tokens} $\h$, one per frame, and the \coltok{object tokens} $\obj_n$, one per four frames, at \colt{diffusion step} $t$. Conditioned on the CLIP \colword{text tokens} $\textok$, the \colword{initial poses}, and the \colword{name} and \colword{BPS} geometry of every object, it generates the human motion and each object motion relative to the joint that drives it.}
    \vspace{-1pt}
    \label{fig:method}
\end{figure}

\methodname{} is a single text-conditioned model for a variable number of objects (\cref{fig:method}): no parameter depends on the object count, so one model generates interactions with one to five objects, articulated and worn ones included, in a single denoising pass.
Given a prompt $\textok$, the initial pose of the person, and a set of $\nobj$ objects, each defined by mesh, name, and initial pose, it generates the human motion $\h$ and the object motions $\obj_1, \dots, \obj_\nobj$ over $\window{=}120$ frames, 4\,s at 30\,fps.

\paragraph{Human representation.}
We represent the person with the gender-neutral SMPL-X body model~\citep{pavlakos2019smplx} with a given shape, as in the \datasetname{}, in a canonical frame defined by the initial pose of the window.
The human modality is
\begin{equation}
    \h = \big\{ \hjoints^t,\, \hpose^t,\, \hhands^t,\, \hcontact^t \big\}_{t=1}^{\window},
    \label{eq:human_modality}
\end{equation}
where, for every frame $t$, $\hjoints^t$ holds global joint positions, $\hpose^t$ the global rotations of the body joints in the 6D representation~\citep{zhou2019continuity}, $\hhands^t$ the PCA coefficients of the finger pose of both hands, and $\hcontact^t$ floor-contact scores of the feet, 256 values per frame in total (\cref{sec:app_human}).

\paragraph{Object representation.}
Each object $\obj_n$ is defined analogously, with the index $n$ omitted:
\begin{equation}
    \obj = \big\{ \objglobal^t,\, \objhinge^t,\, \{\objrel{j}^t\}_{j \in \relj},\, \objcontact^t \big\}_{t=1}^{\window}.
    \label{eq:object_modality}
\end{equation}
The \emph{absolute pose} $\objglobal^t$ is the rotation and the center of mass of the object in the canonical frame of the window, and the \emph{hinge channel} $\objhinge^t$ encodes the articulation angle of objects with a moving part.
The \emph{relative poses} $\objrel{j}^t$ express the same pose in the frames of six candidate driving joints $\relj = \{\text{spine},\text{head},\text{both wrists},\text{both feet}\}$, through \cref{eq:relative}, so that an object held in a hand has a nearly constant pose in the wrist frame whatever the rest of the body does; the body alone is in fact predictive of the object pose~\citep{petrov2023popup}.
Finally, the \emph{contact scores} $\objcontact^t$ are soft indicators of contact between the object and 15 body regions (\cref{sec:app_object}).

\paragraph{Conditioning.}
The text prompt is encoded with CLIP~\citep{radford2021learning} and enters the model both as a pooled sentence embedding and as the sequence of word-piece embeddings, the text tokens $\textok$.
Following HIMO~\citep{lv2024himo}, we also condition on the first frame of the window, the initial pose of the person and of every object.
Each object is further described by its name and by an embedding of its geometry, computed from basis point set features~\citep{prokudin2019efficient} with a hierarchical encoder (\cref{sec:app_geometry}).
Since the object names appear in the captions and are encoded with the same CLIP model, the text tokens, as in MotionDiffuse~\citep{zhang2024motiondiffuse}, let every object be related to the words that refer to it, which the pooled embedding cannot.

\paragraph{Denoiser.}
The denoiser $\methodnetwork$ is a transformer~\citep{vaswani2017attention} with one token per frame of the human motion $\h$ and of every object motion $\obj_n$, all attending to each other and to the conditioning tokens.
Since attention cost grows quadratically with the tokens, we chunk every four frames of an object into one token.
A smoothing convolution decodes the output tokens to per-frame parameters, without which the object motion has a seam every four frames (\cref{sec:app_architecture}).

\paragraph{Object slots.}
The token streams of the objects, which we call slots, receive no slot-specific embedding: all slots share the input and output projections and differ only in their conditioning, and masks exclude empty slots from attention and from the per-object losses.
The network is therefore permutation-equivariant with respect to the objects, and none of its parameters depends on their number, which is set at run time rather than in the architecture, with up to five slots in training.

\paragraph{Part assignment.}
We say that a joint $j \in \relj$ \emph{drives} an object if the object moves with this joint: an object held in a hand, worn on the head or back, or kicked with a foot is driven by the wrist, head, spine, or foot.
A chair, in contrast, is not driven but only supports the person who sits on it.
A lightweight transformer head predicts the driving joints of every object from the conditions alone, supervised with labels derived from contacts and joint motion (\cref{sec:app_assignment}).
Its prediction enters the denoiser as an additional input and selects the decoding: a driven object is placed through the relative pose $\objrel{j}^t$ of its driving joint, and an undriven object keeps its absolute pose $\objglobal^t$.

\paragraph{Objective.}
We use the DDPM formulation~\citep{ho2020denoising} with a cosine noise schedule of 1000 steps~\citep{nichol2021improved}, and the network predicts the clean sample~\citep{ramesh2022hierarchical}.
Let $\vec{x} = (\h, \obj_1, \dots, \obj_\nobj)$ denote a window, $\vec{x}^{\tau}$ its noisy version at diffusion step $\tau$, and $\cond$ the conditions.
The denoiser $\methodnetwork$ is trained with
\begin{equation}
    \mathcal{L} = \mathbb{E}_{\vec{x}, \tau}\; w(\tau) \sum_{k} \lambda_k\, \loss{k}\big(\methodnetwork(\vec{x}^{\tau}; \tau, \cond), \vec{x}\big),
    \label{eq:objective}
\end{equation}
where $w(\tau)$ is the min-SNR weight~\citep{hang2023efficient}.
The terms $\loss{k}$ cover the reconstruction of the parameters and their frame-to-frame differences, the contact scores, and the plausibility of the body through foot-skating and forward-kinematics penalties.
A consistency term further poses points of the object surface with the absolute and with each relative pose, which describe the same motion, and penalizes their distance to the ground truth (all weights $\lambda_k$ in \cref{sec:app_losses}).

\paragraph{Training and inference.}
Windows are cut at random offsets and paired with one of their captions.
We drop the text prompt for classifier-free guidance~\citep{ho2022classifier} and, for robustness to missing inputs, also worn objects, names, and the part-assignment input (\cref{sec:app_training}).
We sample with 100 DDPM steps respaced from the 1000 training steps and a classifier-free guidance scale of 2.5, and generate longer sequences window by window with half-window blended overlaps.

\paragraph{Contact-consistent guidance.}
The model predicts finger-object contacts in $\objcontact^t$ and foot-floor contacts in $\hcontact^t$, but its geometry does not always agree with them.
Since the contact labels decrease with the distance between a finger and the object surface and with the speed of a foot, a predicted contact implies an upper bound on this distance or speed, which holds exactly on the training data.
In the last sampling steps, gradient steps on the predicted clean sample penalize violations of these bounds (\cref{eq:guidance}), which closes the gaps between fingers and objects and stops planted feet.

%% file: sections/05_experiments.tex
\begin{figure}[b]
    \centering
    \vspace{-10pt}
    \includegraphics[width=1.0\linewidth]{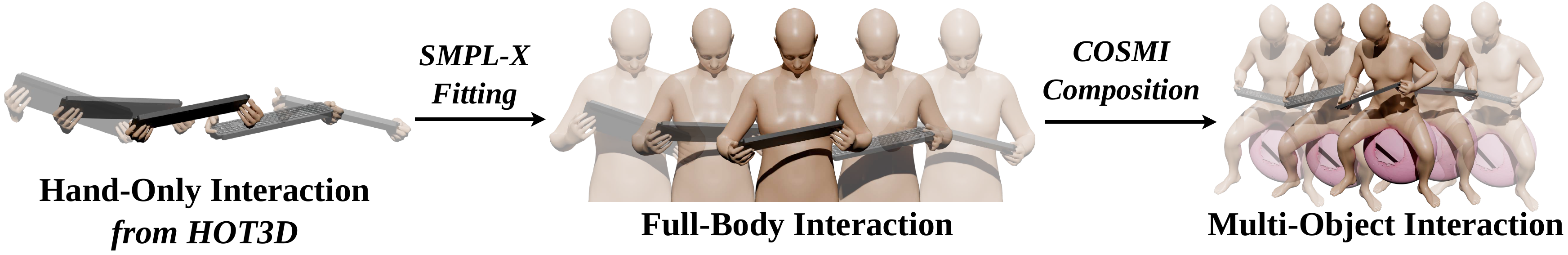}
    \vspace{-20pt}
    \caption{\textbf{Hand-object recordings as source clips.} A HOT3D recording of the hands and the object (left) becomes a full-body clip once SMPL-X is fitted to the headset and the MANO hands (middle), and the composition pairs it with a sitting clip like any captured source (right).}
    \label{fig:hands}
\end{figure}
\begin{figure}[t]
    \centering
    \vspace{-10pt}
    \includegraphics[width=1.0\linewidth]{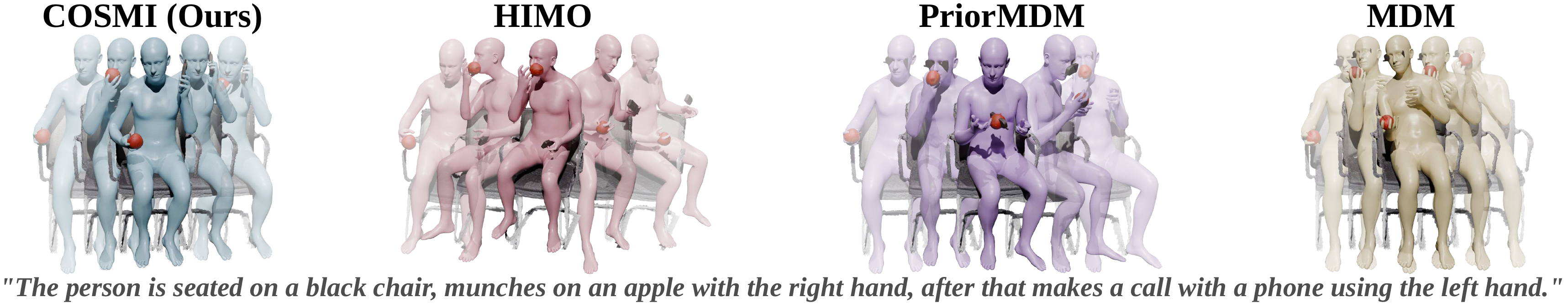}
    \vspace{-20pt}
    \caption{\textbf{Qualitative comparison.} Generations for a test prompt with the unseen black chair and two hand actions in sequence. \methodname{} keeps the person seated, and generates plausible contacts.}
    \label{fig:qualitative}
\end{figure}

We evaluate on our benchmark, compare with HIMO on its dataset, and ablate \methodname{}.
\Cref{fig:hands} shows HOT3D~\citep{banerjee2025hot3d} recordings turned into source clips by fitting a body to the headset and the MANO~\citep{romero2017mano} hands (\cref{sec:app_hot3d}), and \cref{fig:qualitative} compares \methodname{} with the baselines.
Composed sequences also animate clothed scans~\citep{wang20244ddress} registered as SMPL-X+D~\citep{kostyrko2026avaimg}, with held objects re-attached to the hands of the scan (\cref{fig:teaser,fig:supp_scans}), and would drive any SMPL-X-anchored avatar~\citep{eskandar2026dama} the same way.

\paragraph{Data split.}
A split by subject, the common choice for captured datasets, says little about generalization here, because the clips of a held-out subject would still occur in thousands of training sequences.
Instead, we split by composition and hold out what a multi-object model is expected to generalize to.
The \emph{unseen object} is the black chair, held out with all its sequences, a new instance of a seat category whose other members remain in training, so the model has to keep the person seated on it from its geometry, name, and initial pose.
The \emph{unseen combinations} are five pairings whose parts are common in training but never occur together: walking with an apple, sitting with a flip phone, a camera together with a mug, a foot interaction while drinking, and standing with a water bottle.
The held-out sequences form the test set, 35k sequences against 187k for training.

\input{tables/04_5o_eval}

\input{tables/03_2o3o_eval}

\paragraph{Protocol.}
We train a single model on all training sequences and evaluate the same weights throughout.
The \emph{two-object} and \emph{three-object} tiers follow the setting of HIMO~\citep{lv2024himo}, test sequences with exactly two or three rigid interacted objects and no worn objects, and the \emph{all-objects} tier contains every test sequence, with one to five objects, articulated and worn included.
Every test sequence is generated three times over its full length from its caption and initial poses.

\paragraph{Metrics.}
A learned evaluator, a text-motion co-embedding trained on our training split, provides FID~\citep{heusel2017gans}, R-precision (top-1 and top-3), multimodal distance (MM-Dist), and diversity, the metrics of text-to-motion generation~\citep{guo2022humanml3d,lv2024himo}.
Foot sliding and object jerk follow HUMOTO~\citep{lu2025humoto}, penetration counts the frames in which the body enters an object, and all three are measured on the decoded meshes.
Temporal and body contact accuracy (T-Acc, B-Acc)~\citep{peng2026surface} compare the generated body-object contacts with those of the ground truth per frame and per body marker.
For articulated objects, we report the hinge-angle error and the sweep ratio of the generated to the ground-truth hinge motion, and all values are means over three generations (definitions in \cref{sec:app_metrics}).

\paragraph{Baselines.}
The closest prior work is HIMO~\citep{lv2024himo}, whose benchmark also adapts MDM~\citep{tevet2023mdm} and PriorMDM~\citep{shafir2024priormdm} to several objects.
We re-implement MDM and PriorMDM on our representation with our initial-pose conditioning, schedule, and losses, but with a pooled text embedding and a small geometry encoder, so that each is a single model evaluated in all tiers.
MDM has one token per frame, whereas PriorMDM has separate human and object streams with one communication block.
HIMO supports rigid objects only, so its two models are trained on the two- and three-object tiers in its format and scored with our evaluator.
The concurrent SK-HOI~\citep{peng2026surface} is not compared, since its full code is unavailable and its test-time optimization of about four minutes per sequence is out of reach for 35k sequences.

\paragraph{Implementation details.}
\methodname{} has 30.4M parameters, between MDM (20.7M) and PriorMDM (39.6M), and trains in eight hours on one RTX 5090.
A window is generated in 0.16\,s with classifier-free guidance, and in 1.1\,s with the contact-consistent guidance (\cref{sec:app_runtime}).

\paragraph{Comparison with baselines.}
\Cref{tab:comparison_full,fig:qualitative} compare the methods in the all-objects tier, \methodname{} with the contact-consistent guidance of \cref{sec:method}.
\methodname{} obtains the best FID, R-precision, MM-Dist, and foot sliding, and its penetration stays at the ground-truth level, the residual penetration of the captures, which the baselines exceed.
\methodname{} has the lowest hinge-angle error but the smallest hinge sweep, two thirds of the ground truth (\cref{sec:app_failures}).
The two- and three-object tiers agree (\cref{tab:comparison_himo}), the baselines up to seventeen points above the ground-truth penetration with two objects, except that HIMO, trained per tier, has the best top-3 R-precision with three objects.

\paragraph{Comparison on the HIMO dataset.}
To compare with HIMO on captured data, we also re-train \methodname{} on the HIMO dataset and evaluate it with the evaluator of HIMO (\cref{sec:app_himo}).
\methodname{} has the highest top-3 R-precision in both tiers and, with three objects, also the best FID.

\paragraph{Ablations.}
We ablate the guidance, the auxiliary losses, the part assignment, the text tokens and object names, the geometry encoder, and the chunk size, with results in \cref{sec:app_ablations}.
The auxiliary losses matter most, and the part assignment improves both text alignment and contact accuracy.
Text tokens and object names improve text alignment, and the guidance significantly decreases foot sliding and hand-object gaps without damaging the other metrics.

\input{tables/07_generalization}

\paragraph{Generalization.}
\Cref{tab:tiers} reports the contact accuracy on the unseen object and on each unseen combination.
On the combinations, all methods match or exceed their accuracy on the whole test set, with one drop of three points, so they appear to learn the base motion and the hand interaction independently of each other.
\methodname{} is the most accurate on every subset, and the unseen object is harder: the baselines lose four to eight points of temporal contact accuracy on the black chair, whereas \methodname{} stays within one point of its value on the whole test set.

%% file: tables/04_5o_eval.tex
\begin{table}[t]
    \centering
    \caption{\textbf{All-objects tier.} One to five objects per sequence. \methodname{} is closest to the text and to the ground-truth contacts and has the lowest hinge error. Data denotes the ground-truth test sequences, $\h$ and $\obj$ the human and the object metrics. Values are means over three generations, with 95\,\% confidence intervals as superscripts for the learned evaluator. Best in bold, second best underlined, diversity best when closest to the data and the hinge sweep when closest to 1.}
    \label{tab:comparison_full}
    \footnotesize
    \setlength{\tabcolsep}{2pt}
    \begin{adjustbox}{max width=\linewidth}
\begin{tabular}{@{}lccccc@{\hspace{0.8em}}c@{\hspace{0.8em}}c@{\hspace{0.8em}}ccc@{\hspace{0.8em}}cc@{}}
    \toprule
    \multirow{2}{*}{\bf Method}
        & \multicolumn{5}{c@{\hspace{0.8em}}}{\bf Text-to-motion}
        & \multicolumn{1}{c@{\hspace{0.8em}}}{\bf $\h$}
        & \multicolumn{1}{c@{\hspace{0.8em}}}{\bf $\obj$}
        & \multicolumn{3}{c@{\hspace{0.8em}}}{\bf Contact}
        & \multicolumn{2}{c@{}}{\bf Articulated} \\
    \cmidrule(lr){2-6} \cmidrule(lr){7-7} \cmidrule(lr){8-8} \cmidrule(lr){9-11} \cmidrule(lr){12-13}
        & \overmean{FID$\downarrow$}{0.000}\noerr{.000}
        & \overmean{Top-1$\uparrow$}{0.000}\noerr{.000}
        & \overmean{Top-3$\uparrow$}{0.000}\noerr{.000}
        & \overmean{MM-Dist$\downarrow$}{0.000}\noerr{.000}
        & \overmean{Div.$\rightarrow$}{0.00}\noerr{.00}
        & Foot sl.$\downarrow$
        & Jerk$\downarrow$
        & T-Acc$\uparrow$
        & B-Acc$\uparrow$
        & Pen. (\%)$\downarrow$
        & Hinge err.$\downarrow$
        & Sweep$\rightarrow$ \\
    \midrule
    Data
        & \nomean{0.148}\noerr{.000} & 0.740\noerr{.000}        & 0.919\noerr{.000}        & 0.740\noerr{.000}        & 4.31\noerr{.00}
        & 1.879
        & 0.366
        & --             & --             & 34.60
        & --             & -- \\
    \midrule
    MDM
        & 0.148\err{.002}            & 0.619\err{.001}          & 0.861\err{.002}          & 1.089\err{.003}          & 4.11\err{.28}
        & 3.290
        & 0.495
        & 0.901          & 0.934          & 37.49
        & \second{0.547} & \first{0.846} \\
    PriorMDM
        & \second{0.077}\err{.000}   & \second{0.664}\err{.001} & \second{0.876}\err{.001} & \second{0.933}\err{.002} & \first{4.18}\err{.26}
        & \second{1.874}
        & \first{0.390}
        & \second{0.938} & \second{0.936} & \second{35.08}
        & 0.557          & \second{0.707} \\
    \midrule
    \methodname{} (ours)
        & \first{0.049}\err{.000}    & \first{0.696}\err{.003}  & \first{0.900}\err{.001}  & \first{0.827}\err{.002}  & \second{4.15}\err{.32}
        & \first{1.395}
        & \second{0.409}
        & \first{0.980}  & \first{0.968}  & \first{33.96}
        & \first{0.437}  & 0.665 \\
    \bottomrule
\end{tabular}
    \end{adjustbox}
\end{table}

%% file: tables/03_2o3o_eval.tex
\begin{table}[t]
    \centering
    \caption{\textbf{Two- and three-object tiers.} Rigid objects only and no worn objects, the setting of HIMO. Format as in \cref{tab:comparison_full}.}
    \label{tab:comparison_himo}
    \footnotesize
    \setlength{\tabcolsep}{2pt}
    \begin{adjustbox}{max width=\linewidth}
\begin{tabular}{@{}clccccc@{\hspace{0.8em}}c@{\hspace{0.8em}}c@{\hspace{0.8em}}ccc@{}}
    \toprule
    & \multirow{2}{*}{\bf Method}
        & \multicolumn{5}{c@{\hspace{0.8em}}}{\bf Text-to-motion}
        & \multicolumn{1}{c@{\hspace{0.8em}}}{\bf $\h$}
        & \multicolumn{1}{c@{\hspace{0.8em}}}{\bf $\obj$}
        & \multicolumn{3}{c@{}}{\bf Contact} \\
    \cmidrule(lr){3-7} \cmidrule(lr){8-8} \cmidrule(lr){9-9} \cmidrule(lr){10-12}
    &
        & \overmean{FID$\downarrow$}{0.000}\noerr{.000}
        & \overmean{Top-1$\uparrow$}{0.000}\noerr{.000}
        & \overmean{Top-3$\uparrow$}{0.000}\noerr{.000}
        & \overmean{MM-Dist$\downarrow$}{0.000}\noerr{.000}
        & \overmean{Div.$\rightarrow$}{0.00}\noerr{.00}
        & Foot sl.$\downarrow$
        & Jerk$\downarrow$
        & T-Acc$\uparrow$
        & B-Acc$\uparrow$
        & Pen. (\%)$\downarrow$ \\
    \midrule
    \multirow{5}{*}{\rotatebox[origin=c]{90}{2 obj.}}
    & Data
        & \nomean{0.260}\noerr{.000} & 0.510\noerr{.000}        & 0.799\noerr{.000}        & 0.539\noerr{.000}        & 2.91\noerr{.00}
        & 2.478
        & 0.635
        & --             & --             & 11.92 \\
    \cmidrule{2-12}
    & MDM
        & 0.260\err{.005}            & 0.294\err{.003}          & 0.544\err{.005}          & 1.132\err{.003}          & \second{2.99}\err{.18}
        & 6.598
        & 0.863
        & 0.919          & 0.915          & 21.49 \\
    & PriorMDM
        & \second{0.090}\err{.002}   & \second{0.398}\err{.005} & \second{0.692}\err{.002} & \second{0.768}\err{.001} & 3.03\err{.23}
        & \second{2.641}
        & \second{0.570}
        & \second{0.969} & \second{0.930} & \second{16.70} \\
    & HIMO
        & 0.095\err{.003}            & 0.391\err{.003}          & 0.670\err{.002}          & 0.964\err{.002}          & \first{2.96}\err{.17}
        & 5.406
        & 1.450
        & 0.818          & 0.885          & 28.68 \\
    \cmidrule{2-12}
    & \methodname{} (ours)
        & \first{0.066}\err{.002}    & \first{0.405}\err{.007}  & \first{0.699}\err{.003}  & \first{0.737}\err{.007}  & 2.78\err{.21}
        & \first{1.974}
        & \first{0.466}
        & \first{0.993}  & \first{0.962}  & \first{11.94} \\
    \midrule\midrule
    \multirow{5}{*}{\rotatebox[origin=c]{90}{3 obj.}}
    & Data
        & \nomean{0.424}\noerr{.000} & 0.298\noerr{.000}        & 0.580\noerr{.000}        & 1.032\noerr{.000}        & 2.77\noerr{.00}
        & 1.407
        & 0.271
        & --             & --             & 37.15 \\
    \cmidrule{2-12}
    & MDM
        & 0.424\err{.010}            & 0.192\err{.002}          & 0.424\err{.001}          & 1.263\err{.003}          & 2.82\err{.18}
        & \second{1.191}
        & 0.400
        & 0.836          & \second{0.923} & \second{38.62} \\
    & PriorMDM
        & 0.313\err{.003}            & 0.218\err{.001}          & 0.454\err{.002}          & \second{1.204}\err{.003} & \second{2.74}\err{.09}
        & 1.247
        & \first{0.308}
        & \second{0.901} & 0.921          & 39.97 \\
    & HIMO
        & \second{0.284}\err{.003}   & \second{0.243}\err{.002} & \first{0.516}\err{.000}  & 1.233\err{.001}          & 2.84\err{.14}
        & 2.311
        & 0.723
        & 0.808          & 0.917          & 42.85 \\
    \cmidrule{2-12}
    & \methodname{} (ours)
        & \first{0.157}\err{.002}    & \first{0.244}\err{.001}  & \second{0.507}\err{.002} & \first{1.056}\err{.003}  & \first{2.76}\err{.14}
        & \first{0.932}
        & \second{0.399}
        & \first{0.968}  & \first{0.966}  & \first{37.90} \\
    \bottomrule
\end{tabular}
    \end{adjustbox}
\end{table}

%% file: tables/07_generalization.tex
\begin{table}[t]
    \centering
    \caption{\textbf{Unseen object and combinations.} Contact accuracy on the test sequences with the unseen object, the black chair, and with the five unseen combinations.}
    \label{tab:tiers}
    \scriptsize
    \setlength{\tabcolsep}{3pt}
    \resizebox{\linewidth}{!}{%
\begin{tabular}{l|cc|cc|cc|cc|cc|cc}
    \toprule
    \multirow{2}{*}{\bf Method}
        & \multicolumn{2}{c}{\bf black chair}
        & \multicolumn{2}{c}{\bf walk + apple}
        & \multicolumn{2}{c}{\bf sit + flip phone}
        & \multicolumn{2}{c}{\bf camera + mug}
        & \multicolumn{2}{c}{\bf foot + drink}
        & \multicolumn{2}{c}{\bf stand + bottle} \\
        & T-Acc$\uparrow$ & B-Acc$\uparrow$
        & T-Acc$\uparrow$ & B-Acc$\uparrow$
        & T-Acc$\uparrow$ & B-Acc$\uparrow$
        & T-Acc$\uparrow$ & B-Acc$\uparrow$
        & T-Acc$\uparrow$ & B-Acc$\uparrow$
        & T-Acc$\uparrow$ & B-Acc$\uparrow$ \\
    \midrule
    MDM
        & 0.823          & \second{0.917}
        & 0.971          & 0.946
        & 0.903          & 0.930
        & 0.955          & 0.953
        & 0.926          & \second{0.958}
        & \second{0.977} & 0.929 \\
    PriorMDM
        & \second{0.895} & 0.916
        & \second{0.983} & \second{0.950}
        & \second{0.941} & \second{0.937}
        & \second{0.965} & \second{0.956}
        & \second{0.942} & 0.954
        & 0.969          & \second{0.939} \\
    \midrule
    \methodname{} (ours)
        & \first{0.975}  & \first{0.960}
        & \first{0.995}  & \first{0.974}
        & \first{0.981}  & \first{0.959}
        & \first{0.984}  & \first{0.984}
        & \first{0.953}  & \first{0.981}
        & \first{0.993}  & \first{0.967} \\
    \bottomrule
\end{tabular}
}
\end{table}

%% file: sections/06_conclusions.tex
Everyday activities combine objects in countless ways, from drinking while sitting to eating an apple while walking, and the combinations to capture grow faster than any recording effort.
The \datasetname{} shows that they need not all be recorded: interactions are local, so a contact-consistent clip of one interaction carries its limb and object onto a body that performs another.
A language model and geometric checks admit only the combinations that a person would plausibly perform and that remain physically valid, so what emerges is not stitched clips but semantically meaningful, realistic multi-object interactions.
The result is 222k sequences and 275 hours with up to five objects, balanced between the hands by mirroring and nearly as plausible as their captures.
Nothing in the composition is tied to a source dataset, so every further capture multiplies the combinations, full-body datasets through a rule and hand-object recordings once a body is fitted to them, as HOT3D shows.
The \methodname{} method, a text-driven diffusion transformer grounding the objects to the body parts that move them, generates a variable number of objects end to end.
The baselines and \methodname{} alike generalize to combinations never seen in training, and \methodname{} also to an unseen object.

\paragraph{Limitations and future work.}
The composition covers sustained single interactions at once or in sequence.
Extending it to multi-stage activities such as cooking or assembling, whose steps and objects depend on each other, and bringing the data together with interactable scenes, where a cup is taken from a shelf on the way, are the directions we consider most promising.
On the method side, the guidance closes the gaps between fingers and objects, but the finger motion has yet to reach the precision of dedicated grasp synthesis.
Finally, text is a coarse interface: the prompt fixes the objects and actions, but not the path of the person or the timing of sequential actions.

%% file: sections/07_statements_arxiv.tex
\subsection*{Acknowledgments}
Special thanks to Riccardo Marin and Nikita Kister for the helpful discussions.
We thank Margaret Kostyrko and Yuxuan Xue for sharing the SMPL-X registrations that were used for the visualizations.
This work is funded by the Deutsche Forschungsgemeinschaft - 409792180 (EmmyNoether Programme, project: Real Virtual Humans).
Daniel Eskandar is supported by the Konrad Zuse School of Excellence in Learning and Intelligent Systems (ELIZA) through the DAAD programme Konrad Zuse School of Excellence in Artificial Intelligence, sponsored by the Federal Ministry of Education and Research.
The authors thank the International Max Planck Research School for Intelligent Systems (IMPRS-IS) for supporting I.~A.~Petrov.
G. Pons-Moll is a member of the Machine Learning Cluster of Excellence, EXC number 2064/1 – Project number 390727645. 
The project was made possible by funding from the Carl Zeiss Foundation.

\subsection*{Contribution statement}
DE and IP contributed equally as joint first authors, and each may list their name first on their CV and website.
IP co-initiated the core idea, organized the project, led the implementation of the \methodname{} method, co-developed the \datasetname{} pipeline, led the writing, and co-created the figures and visualizations.
DE co-initiated the core idea, led the implementation of the \datasetname{} pipeline, co-developed the \methodname{} method, co-wrote the text, and led the creation of the figures and visualizations.

%% file: sections/08_appendix.tex
\section*{Appendix}
\begin{quotation}
\noindent\textbf{Abstract.}
This appendix extends the experiments of the main paper and completes the description of the dataset and the method.
In \cref{sec:app_results}, we measure the quality of the composed sequences against their parent recordings, ablate the components of the \methodname{} method, compare with HIMO on its own dataset, and present qualitative results and failure cases.
We also show that hand-object recordings enter the composition once a body is fitted to them, and we render composed sequences on clothed scans.
In \cref{sec:app_metrics}, we define the evaluator behind the learned metrics and the physical metrics with their thresholds and units.
In \cref{sec:app_dataset}, we detail the construction of the dataset, from the preprocessing of the sources and the interaction rules to the mirroring that balances the hands and the feasibility checks of the composition.
Finally, in \cref{sec:app_method}, we describe the \methodname{} method in full, including the representations of the person and the objects, the architecture, the training objective, and the contact-consistent guidance.
\end{quotation}

\appendixtoc
\clearpage

\section{Additional Results}
\label{sec:app_results}

\input{tables/02_dataset_metrics_parents}

\subsection{Quality of composed sequences}
\label{sec:app_quality}
\Cref{tab:dataset-metrics-parents} compares, for each of the six sources, the parent recordings with the composed sequences built from them, with the metrics of HUMOTO~\citep{lu2025humoto}: foot sliding and jerk of the human, jerk of the objects, and the depth to which objects penetrate the body and the hands.
Foot sliding is unchanged, since composition alters neither the lower body nor the global trajectory.
Human jerk rises by a fifth to nearly a half on GRAB, ARCTIC, InterCap, and AMASS, most in absolute terms for the walking sequences of AMASS, which carry the arm motion of a manipulation after composition, and object jerk and penetration change in both directions and worsen by less than the captured datasets differ among themselves, the largest relative rise being the body penetration of InterCap, from 0.17 to 0.40\,cm. Objects on walking bodies have no captured counterpart and show the highest object jerk, since they move with the gait.

\afterpage{\input{tables/05_ablations}\input{tables/08_himo_dataset}}

\subsection{Ablations}
\label{sec:app_ablations}
\Cref{tab:ablations} changes one component of the model at a time.
Guidance reduces foot sliding by a fifth and closes the finger-object gaps: on 512 test windows, the share of predicted contacts that violate their bound (\cref{eq:guidance_bounds}) by more than 5\,mm drops from 18.9\,\% to 0.9\,\%.
Text alignment and contact accuracy stay unchanged, since the guidance enforces the predicted contacts rather than correcting them, at the cost of slightly more object jerk and penetration.
The other variants are sampled without guidance and compare with the unguided model.
The auxiliary losses matter most: without them, foot sliding doubles and temporal contact accuracy drops by 17 points.
The part assignment improves both text alignment and contacts, whereas the text tokens and the object names improve the text alignment only, by two and one points of top-1 R-precision.
Replacing the hierarchical geometry encoder by an MLP changes little, slightly worse top-3 R-precision and multimodal distance against a better FID and object jerk, and the chunk size of four balances single-frame object tokens, which lower the temporal contact accuracy by two points, against chunks of eight, which lower the top-1 R-precision by two points.

\subsection{Comparison on the HIMO dataset}
\label{sec:app_himo}
\Cref{tab:himo_dataset} compares \methodname{}, trained on the HIMO dataset and sampled without guidance, the setting closest to that of HIMO, with the values that HIMO reports for MDM, PriorMDM, and HIMO under its evaluator.
\methodname{} has the highest top-3 R-precision in both tiers, 0.730 and 0.582 against 0.637 and 0.535 for HIMO, and with three objects also the lowest FID and MM-Dist and the diversity closest to the data.
With two objects, HIMO, trained for this number of objects alone, has a lower FID, 1.481 against 4.896, and a lower MM-Dist, and our multimodality is the lowest in both tiers.

\subsection{Held-out subsets}
\label{sec:app_splits}
The split assigns each test sequence to the first matching subset: black chair 12,883 sequences; walking with an apple 10,737; sitting with a flip phone 4,630; a foot interaction while drinking 3,051; standing with a water bottle 2,018; a camera together with a mug 1,799.
The remaining 186,559 sequences form the training set.

\subsection{Qualitative comparison}
\label{sec:app_qualitative}
\begin{figure}[p]
    \centering
    \includegraphics[width=\linewidth,height=0.9\textheight,keepaspectratio]{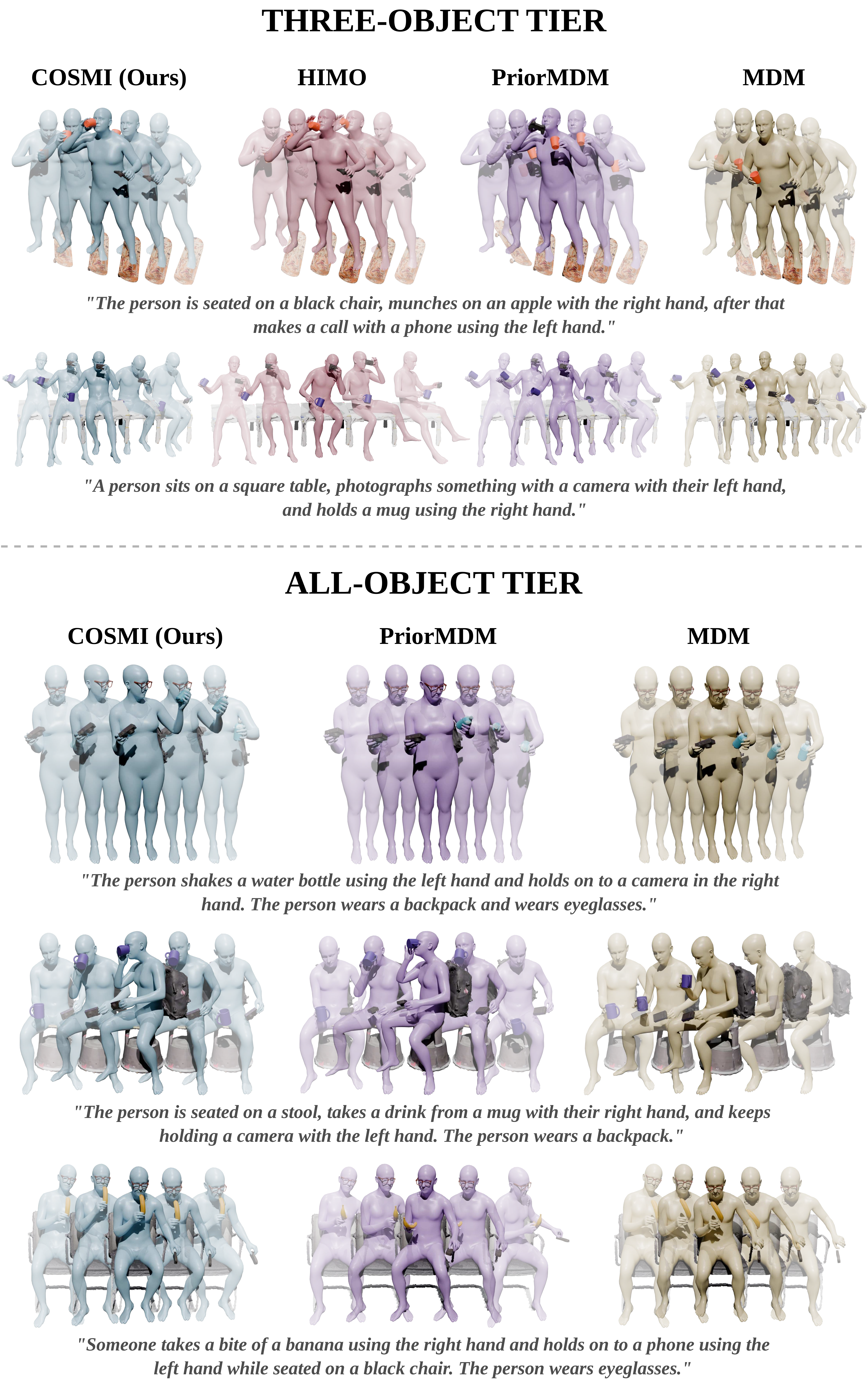}
    \caption{\textbf{Qualitative comparison with the baselines.} Generations for test prompts of the three-object tier (top) and of the all-objects tier (bottom), which includes more than 3 objects and therefore no HIMO. \methodname{} keeps every object in the hand that drives it, whereas the baselines let objects drift away from the hands or float beside the body.}
    \label{fig:supp_results}
\end{figure}

\Cref{fig:supp_results} compares \methodname{} with the baselines on test prompts of the three-object and the all-objects tier, all from the unseen object or combinations.
\methodname{} keeps every object in the hand that drives it over the whole sequence, an apple at the mouth or a camera in front of the face, whereas the baselines, which have no part assignment and attach an object to a joint only while its predicted contact score stays high, let them drift away from the hand or float beside the body once the arm moves.
HIMO appears in the three-object tier only, as it only supports a fixed number of objects.

\subsection{Failure cases}
\label{sec:app_failures}
\begin{figure}[t]
    \centering
    \includegraphics[width=\linewidth]{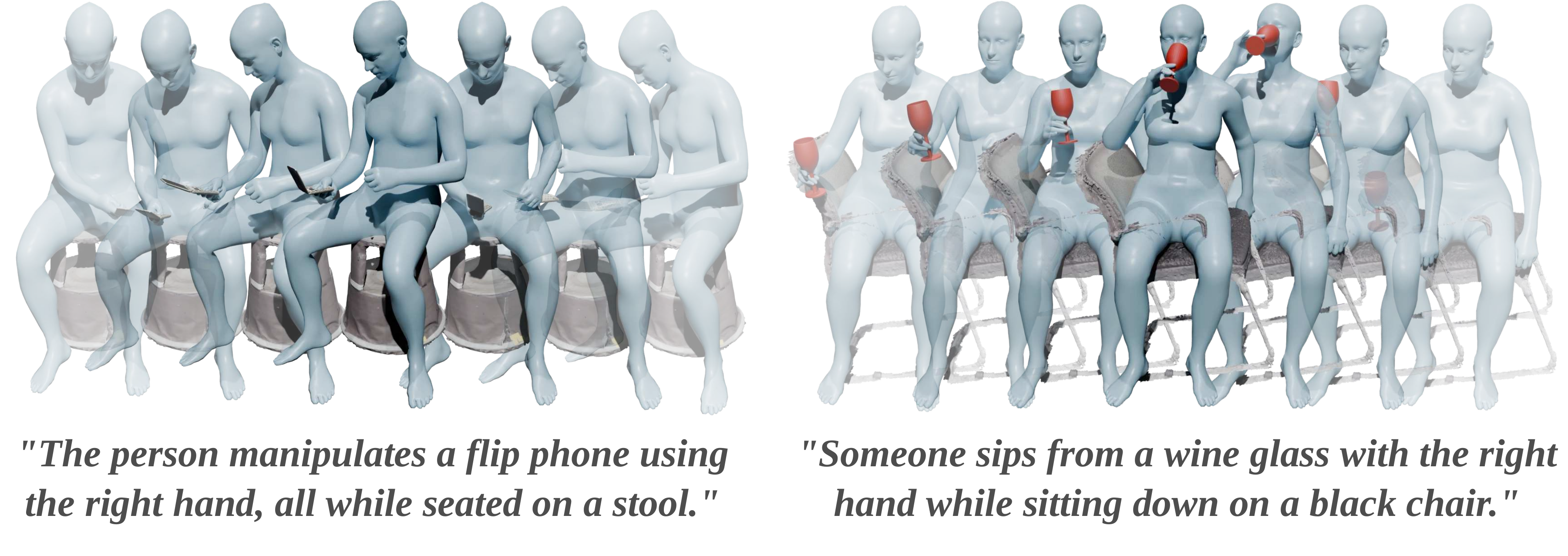}
    \caption{\textbf{Failure cases.} Left: the hinge of the flip phone has limited movement range. Right: the unseen black chair is placed off the body, which intersects its armrest.}
    \label{fig:failures}
\end{figure}

\Cref{fig:failures} shows the two failure modes of \methodname{}.
Articulated objects are the first: the hinge moves less than in the ground truth, a sweep ratio of 0.665 in \cref{tab:comparison_full}, and the moving part is sometimes off its true angle.
The second is the unseen black chair, which is occasionally placed incorrectly relative to the body.
The seats seen in training, such as the wooden chair, the yoga ball and the stool, have no armrests, so the model has no example of keeping the arms clear of them.

\subsection{Hand-object recordings}
\label{sec:app_hot3d}
\begin{figure}[p]
    \centering
    \includegraphics[width=\linewidth,height=0.9\textheight,keepaspectratio]{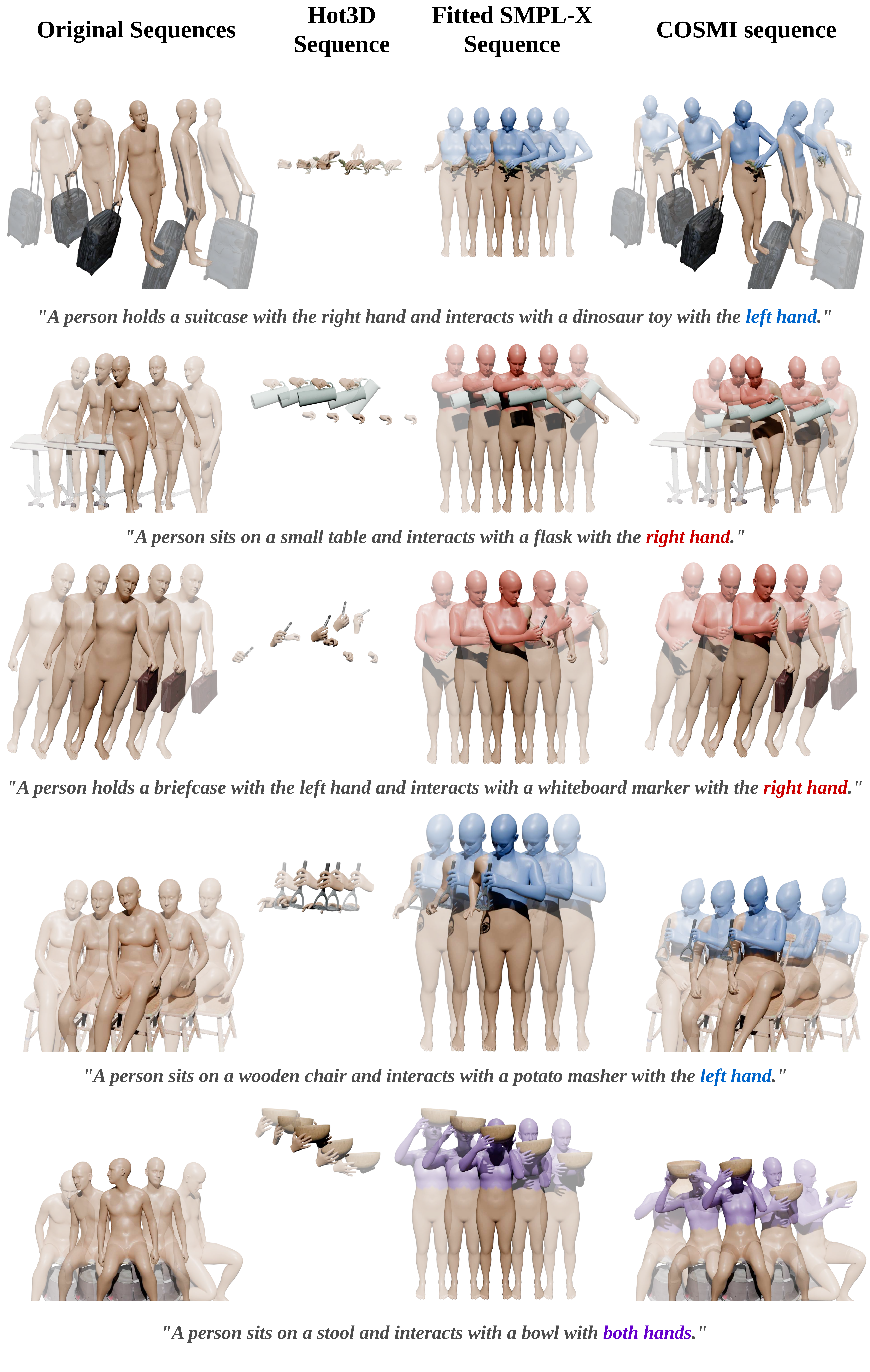}
    \caption{\textbf{Hand-object recordings as source clips.} A HOT3D recording (second column) provides the hands and the object. A body fitted to the head-mounted device and the MANO hands (third column) turns it into a source clip, which the rules and the composition process like any other, here onto sitting and large-object targets (first column), giving the composed sequences (fourth column).}
    \label{fig:supp_hands}
\end{figure}

HOT3D~\citep{banerjee2025hot3d} provides the trajectory of the head-mounted device, MANO~\citep{romero2017mano} hand poses, and object poses.
For every object, we keep the segments in which the head is stable and exactly one object is lifted and in contact with a hand, and place a neutral SMPL-X template with its head at the tracked head position, facing the object.
The rotations of the upper body are then optimized so that the SMPL-X hand vertices match the MANO hands and the head faces the object, with a pose prior and a temporal smoothness term, and the finger poses are copied from MANO.
The resulting clips contain a full body with the recorded hands and object, and the interaction rules process them like any other source clip, so HOT3D objects can be composed onto the walking, sitting, and large-object targets of the dataset (\cref{fig:supp_hands}).

\subsection{Clothed scans}
\label{sec:app_scans}
\begin{figure}[p]
    \centering
    \includegraphics[width=\linewidth,height=0.9\textheight,keepaspectratio]{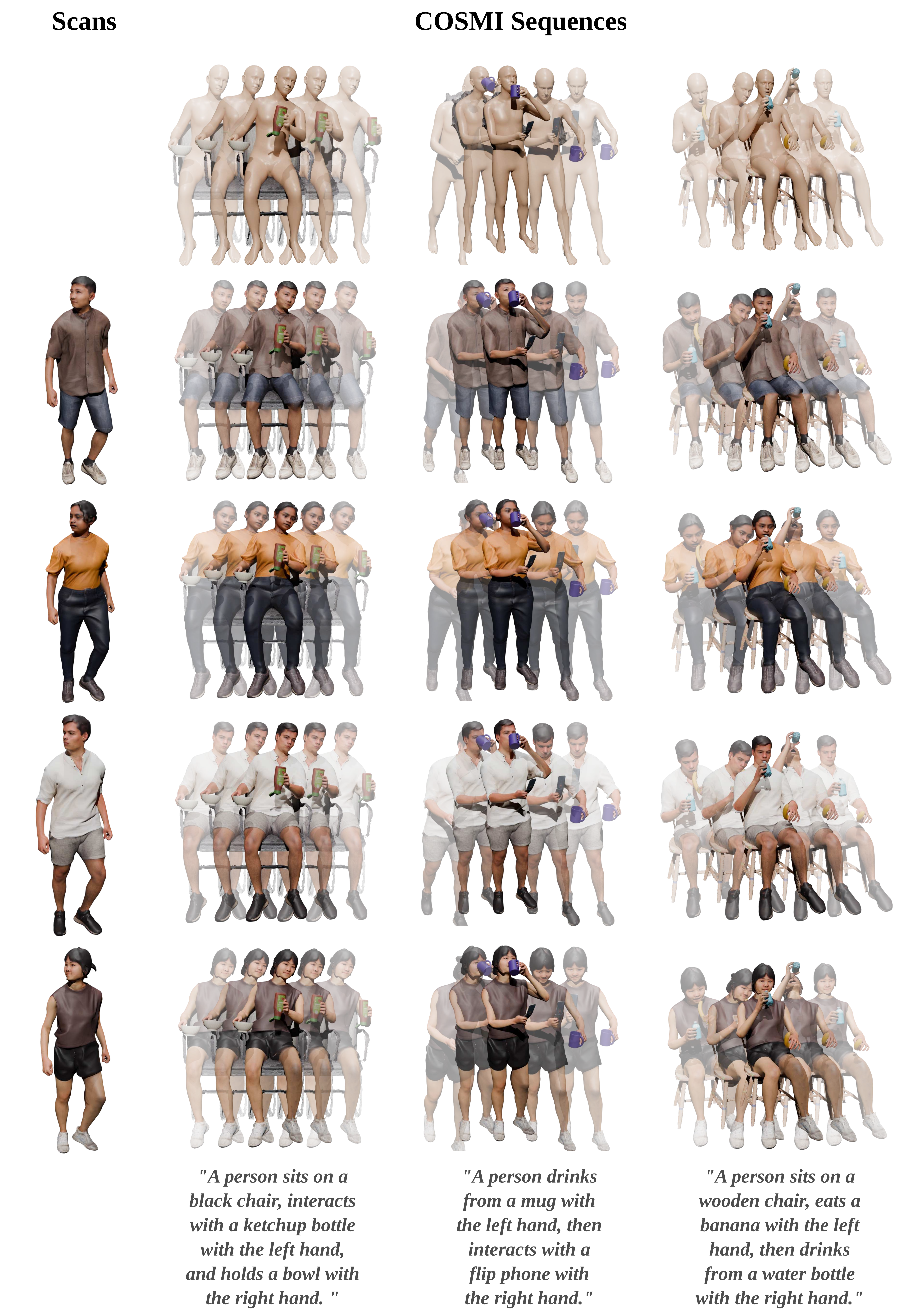}
    \caption{\textbf{Clothed scans animated with composed sequences.} 4D-DRESS scans (left), registered as SMPL-X+D, are driven by the joint transforms of composed sequences, shown on the body model in the top row. Objects follow through their joint-relative transforms, and the finger motion is preserved on the hands.}
    \label{fig:supp_scans}
\end{figure}

We use the SMPL-X+D registrations from~\citep{kostyrko2026avaimg} of the 4D-DRESS scans~\citep{wang20244ddress}, per-vertex displacements on top of the body model, which we animate through the SMPL-X skinning weights with the joint transforms of the composed sequence.
The displacements are attenuated along the forearms and removed on the hands, so that the recorded finger motion is preserved, and the objects are re-attached through their joint-relative transforms, so that a held object follows the hand of the scan (\cref{fig:teaser,fig:supp_scans}).
All figures and videos are rendered with Blendify~\citep{blendify2024}.

\section{Metric Definitions}
\label{sec:app_metrics}

\subsection{Evaluator}
\label{sec:app_evaluator}
The learned metrics follow the text-to-motion protocol~\citep{guo2022humanml3d} with an evaluator trained on our training split.
A sequence is described per frame by the 24 joint positions and, per object slot, the 6D rotation, the center of mass, the hinge channel, and a presence flag, z-scored with the training statistics.
A convolutional movement encoder, trained first as an autoencoder on 64-frame snippets and downsampling by four, feeds a bidirectional GRU motion encoder.
The motion encoder is trained together with a projection of the frozen CLIP text embedding into a common 512-dimensional space with a contrastive margin loss, margin 3, where pairs with identical canonical captions are never negatives.
R-precision and MM-Dist are computed on pools of 32 sequences, FID between the embeddings of the generations and of the ground-truth sequences, and diversity over 100 random pairs, each averaged over the three generations and reported with its 95\,\% confidence interval.
As these metrics depend on the evaluator, we compare them only between methods scored with the same evaluator.
Multimodality (MModal.), reported on the HIMO dataset only, is the mean distance between the embeddings of generations for the same caption, as in HIMO.

\subsection{Physical metrics}
\label{sec:app_physical}
Foot sliding~\citep{li2024chois,lu2025humoto} is the horizontal displacement of the ankle and toe joints in frames where they are below a height threshold $H$, 8\,cm for the ankles and 4\,cm for the toes, which is one on the floor and zero at the threshold, summed, divided by the number of frames, and averaged over the four joints, in mm per frame.
Object jerk is the mean norm of the third temporal difference of the object position, as in HUMOTO~\citep{lu2025humoto}, in cm per frame$^3$.
Penetration is the fraction of frames in which a body vertex lies deeper than 2\,cm inside an object, the collision metric of SK-HOI~\citep{peng2026surface}.
\Cref{tab:dataset-metrics-parents} instead uses the definitions of HUMOTO, whose penetration is the deepest intrusion of the body or hands into the object, averaged over frames.
For the contact accuracy~\citep{peng2026surface}, a body marker is in contact when it lies within 2\,cm of the object surface: T-Acc is the accuracy of the per-frame contact state, where a frame is in contact when any of the 138 markers is, and B-Acc the accuracy of the per-marker states.
The hinge-angle error is the mean absolute difference between the generated and the ground-truth hinge angle, wrapped to $[-\pi, \pi]$, in radians, and the sweep ratio divides the total hinge motion of a generated sequence by that of the ground truth.
The 95\,\% confidence intervals of these metrics are below 0.02, and below 0.2 percentage points for penetration.

\clearpage

\section{Dataset Details}
\label{sec:app_dataset}

\subsection{Preprocessing}
\label{sec:app_preprocessing}
BEHAVE and InterCap enter in their InterAct~\citep{xu2025interact} versions, which correct jitter, contacts, and hand poses, and BEHAVE is converted from SMPL+H to SMPL-X.
The subject shapes of GRAB and ARCTIC are fitted to the gender-neutral model with SMPLFitter~\citep{sarandi2024nlf}, so that every source uses neutral shape parameters.
All sequences are resampled to 30\,fps and expressed in a Y-up frame with the floor at zero and the body centered on the ground plane in the first frame.
Every object is a template mesh with a rotation and a translation per frame.
The labels that the annotation corrects are the InterAct descriptions of GRAB, BEHAVE, and InterCap, the descriptions of OMOMO, a template sentence for ARCTIC, and the BABEL labels of the walking sequences.

\subsection{Interaction rules}
\label{sec:app_rules}
\begin{figure}[p]
    \centering
    \includegraphics[width=\linewidth,height=0.9\textheight,keepaspectratio]{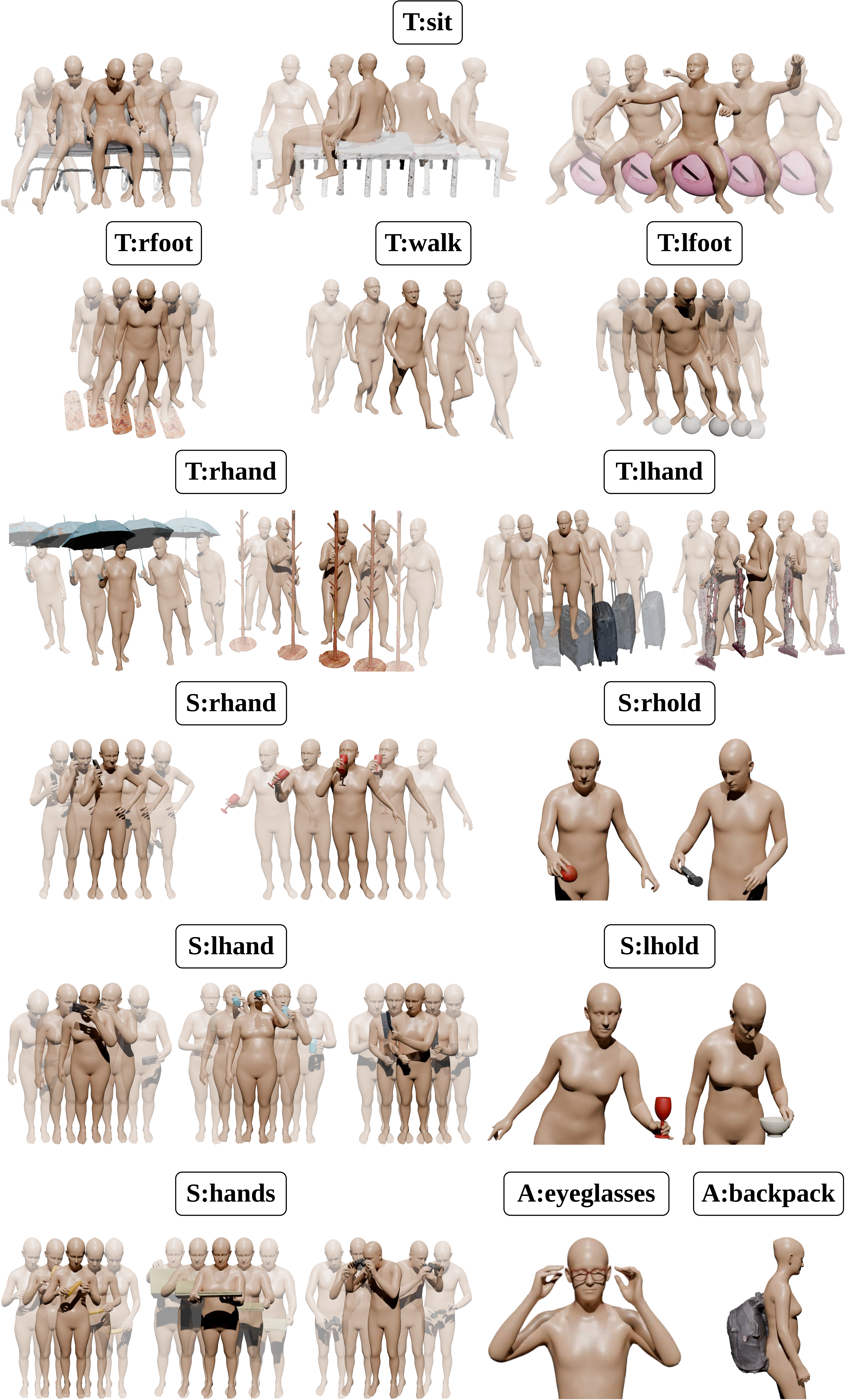}
    \caption{\textbf{Atomic clips per interaction rule.} Clips extracted by each of the thirteen rules of \cref{tab:rules}, the walking sequences of AMASS included. Target rules (T) engage the whole body, source rules (S) one or both hands, and augmentation rules (A) worn objects. Holds and worn objects are single frames.}
    \label{fig:supp_rules}
\end{figure}

\Cref{tab:rules} lists the thirteen rules with their contact groups, pose constraints, local references, and datasets, together with the number of clips each yields before and after mirroring, and \cref{fig:supp_rules} shows examples for every rule.
A rule holds in a frame when every part of its contact group is within 5\,cm of the object surface (\cref{sec:dataset_extraction}), no part of its no-contact group is, and its pose constraint is met.
An upright torso, the most common constraint, is a pelvis-to-head direction within 20 to 30 degrees of the vertical, depending on the rule.

\input{tables/10_rules}

\subsection{Annotation prompts}
\label{sec:app_prompts}
Qwen3 receives the rule description, the object name, the given label, and the narrative of the source dataset, and answers in two steps (\cref{fig:prompts}).
It first picks the verb that names the sustained interaction from a fixed inventory of 28 verbs, \emph{hold, drink, move, lift, carry, push, pull, drag, switch on, take a picture, play, rotate, kick, dribble, pass, call, eat, see, browse, swing, manipulate, pour, toast, use, shake, clean, set down, interact}.
It then writes one sentence with that verb, the object, and the acting hand in a fixed form, which the composition later parses.
Whether an action requires gaze at the object and involves the head is asked once per verb and object pair in a third query of the same form, and the answer is shared by all clips with that pair.

\begin{figure}[p]
    \footnotesize
    \tcbset{prompt/.style={enhanced, colback=gray!8, colframe=black!75, colbacktitle=black!75, coltitle=white, fonttitle=\footnotesize\bfseries, top=2pt, bottom=2pt, left=6pt, right=6pt, toptitle=2pt, bottomtitle=2pt, colbacklower=gray!20, segmentation style={solid, black!40}}}
    \begin{tcolorbox}[prompt, title={Action correction}]
    {\itshape Pick the single verb (from the list below) that best names the main, sustained interaction with ``briefcase'' in this narrative.}

    \textbf{Rule:} The right hand is in contact with the object while the left hand is not.\\
    \textbf{Object:} briefcase\\
    \textbf{Given label (may be wrong):} \texttt{hold}\\
    \textbf{Narrative:} A person holds a briefcase with their right hand, extends their right arm to move the briefcase from their right side to their left front, while taking a step to the left with their left foot.

    \textbf{Allowed verbs} (choose exactly one, copied as written):\\
    \texttt{hold, drink, move, lift, carry, push, pull, drag, switch on, take a picture, play, rotate, kick, dribble, pass, call, eat, see, browse, swing, manipulate, pour, toast, use, shake, clean, set down, interact}

    \textbf{Guidelines:}
    \begin{itemize}[noitemsep, topsep=1pt, leftmargin=1.2em]
        \item The verb must describe something happening TO the object, not the person's general body movement while the grip stays fixed.
        \item \texttt{carry} = walking while holding; \texttt{push}/\texttt{pull}/\texttt{drag} = sliding without lifting.
        \item \texttt{move} = fallback when displaced but none of the above fit.
        \item Ignore setup/teardown (initial grab, final put-down).
    \end{itemize}
    \textit{Reply with only the chosen verb.}
    \tcblower
    \textbf{Qwen3:}\quad\texttt{move}
    \end{tcolorbox}
    \begin{tcolorbox}[prompt, title={Sentence generation}]
    {\itshape Write one short sentence describing a person's interaction with an object.}

    \textbf{Rule:} The right hand is in contact with the object while the left hand is not.\\
    \textbf{Object:} briefcase\\
    \textbf{Action} (use exactly, do not swap for a synonym): \texttt{move}

    \textbf{Requirements:}
    \begin{itemize}[noitemsep, topsep=1pt, leftmargin=1.2em]
        \item Conjugate \texttt{move} for subject ``A person''; pick whatever preposition sounds natural.
        \item Present tense, one action, no direction or speed, no sub-parts.
        \item Refer to the object as ``a briefcase''.
        \item Start with ``A person'', end with ``with the right hand'', 4--10 words.
    \end{itemize}
    \textbf{Examples:}
    \begin{itemize}[noitemsep, topsep=1pt, leftmargin=1.2em]
        \item A person pours from a water bottle with the right hand.
        \item A person drinks from a cup with the right hand.
        \item A person eats an apple with the right hand.
        \item A person calls with a phone with the right hand.
        \item A person passes a game controller with the right hand.
    \end{itemize}
    \textit{Reply with only the sentence.}
    \tcblower
    \textbf{Qwen3:}\quad\texttt{A person moves a briefcase with the right hand.}
    \end{tcolorbox}
    \vspace{-10pt}
    \caption{\textbf{Annotation prompts.} The two per-clip prompts of the annotation with the answer of Qwen3, for a clip of the InterCap briefcase whose given label \emph{hold} is corrected to \emph{move}.}
    \label{fig:prompts}
\end{figure}

\subsection{Joint-relative objects}
\label{sec:app_relative}
The global transform of an object is recovered from its relative transform (\cref{eq:relative}) and the world transform of the reference joint of any body, $\mat{R}^t_\obj = \mat{R}^t_j \mat{R}^t_{\obj|j}$ and $\vec{t}^t_\obj = \mat{R}^t_j \vec{t}^t_{\obj|j} + \vec{t}^t_j$.
When the local reference holds several joints, as for two-handed manipulation, one relative transform per joint is stored, the global transform is recovered from every joint, and the results are averaged, with rotations averaged as unit quaternions and renormalized.

\begin{figure}[tp]
    \centering
    \includegraphics[width=0.95\linewidth]{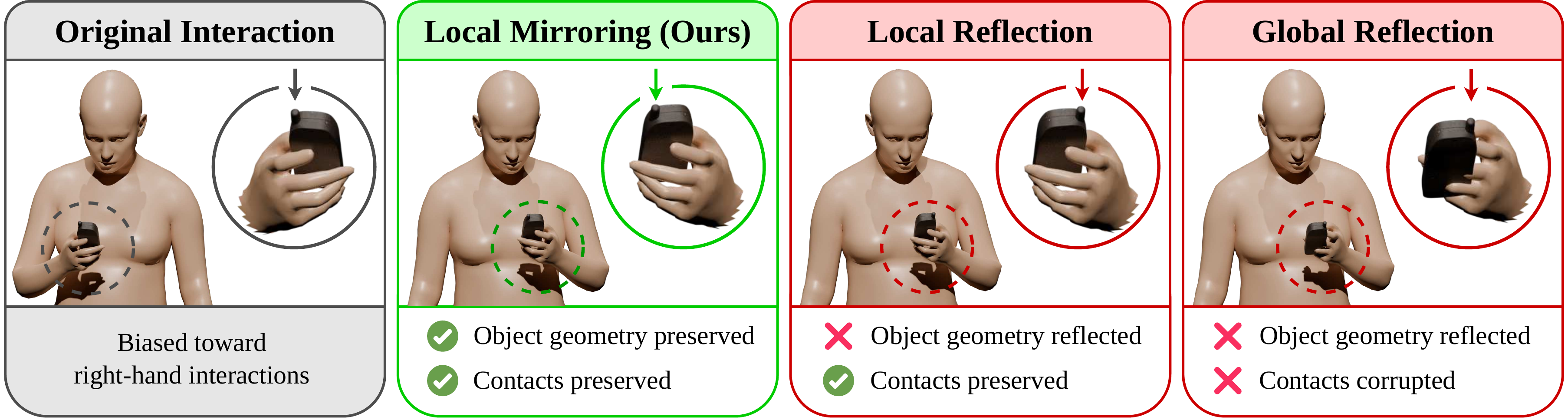}
    \vspace{-10pt}
    \caption{\textbf{Mirroring.} Global reflection of the object breaks the contact with the mirrored hand, and local reflection keeps the contact but mirrors the mesh. Our local mirroring poses the original mesh like its reflection with a proper rotation and preserves both.}
    \label{fig:mirroring}
\end{figure}

\subsection{Mirroring}
\label{sec:app_mirroring}
\Cref{fig:mirroring} compares three ways of mirroring the object of a clip whose body has been reflected with $\mir$ and had its left and right joints swapped.
Reflecting the world pose of the object breaks the contact, because SMPL-X is not symmetric and the mirrored hand is not exactly where the reflection puts the object.
Reflecting the joint-relative pose keeps the contact but poses the mirror image of the mesh: a reflected posed object, $\mir(\mat{R}^t_{\obj|j} \mat{V}_\obj + \vec{t}^t_{\obj|j}) = (\mir \mat{R}^t_{\obj|j} \mir)(\mir \mat{V}_\obj) + \mir \vec{t}^t_{\obj|j}$, is the reflected template $\mir \mat{V}_\obj$ under a proper rotation and a translation.
Our mirroring keeps the original template and uses the alignment of \cref{eq:mirror_align}, which says that $\mir \mat{V}_\obj$ is, up to the residual, the original template rotated by $\mat{Q}_\obj$ about its centroid and shifted by $\vec{p}_\obj$.
Substituting this for $\mir \mat{V}_\obj$ gives the mirrored relative transform
\begin{equation}
    \tilde{\mat{R}}^t_{\obj|j} = \mir \mat{R}^t_{\obj|j} \mir \mat{Q}_\obj, \qquad
    \tilde{\vec{t}}^t_{\obj|j} = \mir \vec{t}^t_{\obj|j} + \mir \mat{R}^t_{\obj|j} \big[ (\mat{I} - \mir \mat{Q}_\obj)\, \bar{\vec{v}}_\obj + \mir \vec{p}_\obj \big],
    \label{eq:mirror_relative}
\end{equation}
a proper rotation and a translation of the original template, so the mesh stays the same and the contact with the mirrored hand is preserved.
The local reference is swapped to the opposite side, the global transform is recovered from the mirrored body, and the rule is swapped accordingly.
The alignment of \cref{eq:mirror_align} is exact for symmetric objects.
Over the 26 mirrored templates, articulated parts counted separately, the distance between the reflected and the aligned template averages 0.9\,mm.
The largest residuals belong to the largest objects, the upper part of the vacuum cleaner (4.0\,mm), the umbrella (3.9\,mm), and the skateboard (2.1\,mm), which have no mirror symmetry.
The fit then favors the regions that matter for the contact, the handle of the umbrella and the deck of the skateboard, at the cost of the rest.
Relative to the bounding-box diagonal, every residual stays below 0.7\,\%.

\begin{wrapfigure}{r}{0.5\linewidth}
    \vspace{-12pt}
    \centering
    \includegraphics[width=\linewidth]{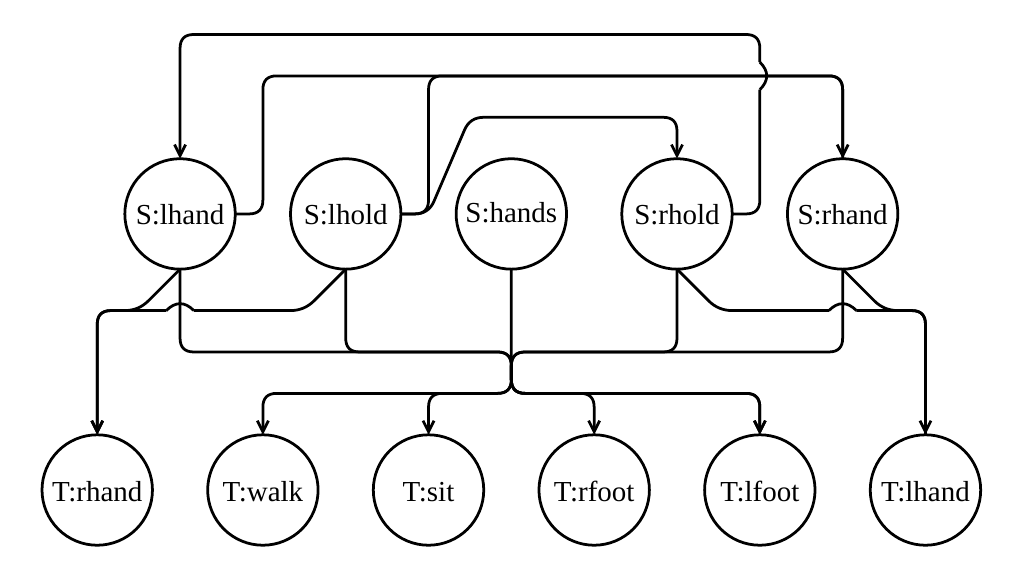}
    \vspace{-6pt}
    \caption{\textbf{Composition graph.} An arrow leads from every source rule to the rules it can be composed onto.}
    \label{fig:graph}
    \vspace{-8pt}
\end{wrapfigure}
\subsection{Composition graph}
\label{sec:app_graph}
\Cref{fig:graph} shows which source rules each rule accepts when it serves as the target: whole-body targets accept every source, a one-hand target accepts the action and the hold of the other hand, and among sources a hand action or hold accepts the other hand, each unordered pair once.
Pairs of the same role or with overlapping limbs are excluded by construction, and the remaining pairs pass through the semantic and physical checks of \cref{sec:dataset_composition}.

\subsection{Merge and feasibility}
\label{sec:app_merge}
The composed pose is the target pose with the joints of the source action replaced: the wrist, elbow, shoulder, and collar of the acting arm, or both arms for a two-handed action.
Every manipulation also replaces the two upper spine joints, because the torso bends with a reaching arm, whereas a passive hold replaces the arm alone.
An action that requires gaze adds the neck and head, one that involves the head adds the spine as well, and an object the target already holds moves with its wrist when the spine changes.
The source objects and worn objects then follow their reference joints through \cref{eq:relative} (\cref{fig:supp_composition}).
The composition is kept if every object retains the contacts of its rule in at least half of the frames.
It is rejected if, in any frame, an object comes within 1\,cm of another object or enters the lower body, the hands enter the lower body, or the arms cross.
Walking targets, whose arms swing, tolerate such self-intersections in up to a third of the frames.

\subsection{Sequential composition and levels}
\label{sec:app_sequential}
\begin{figure}[p]
    \centering
    \includegraphics[width=\linewidth,height=0.9\textheight,keepaspectratio]{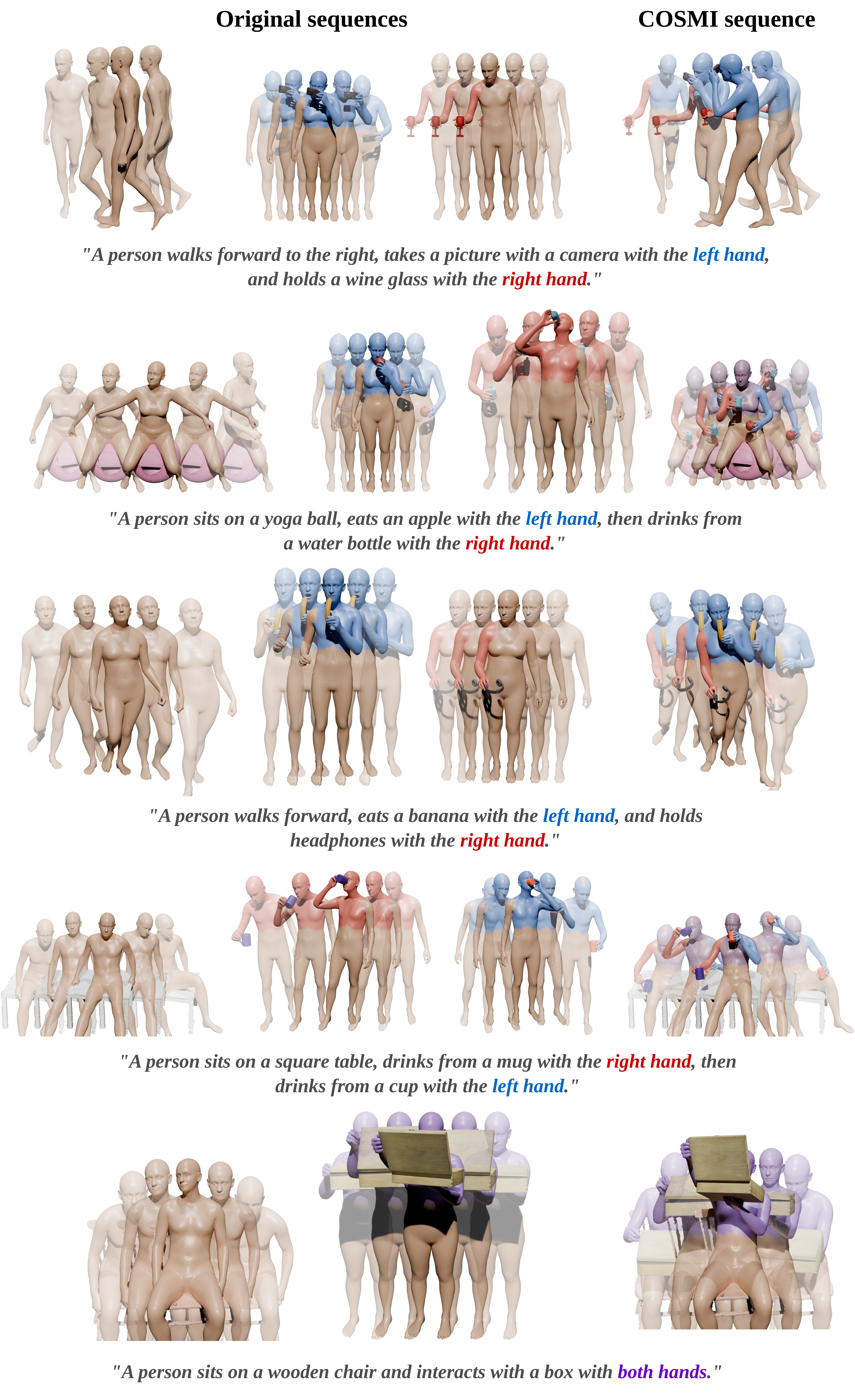}
    \caption{\textbf{Composition of sequences.} Left: the target clip, which provides the body motion, and the source clips, whose limbs are colored, blue for the left hand, red for the right hand, and purple for both hands. Right: the composed sequence, in which the source limbs replace their counterparts and every object follows its limb through \cref{eq:relative}. Rows two and four are sequential compositions.}
    \label{fig:supp_composition}
\end{figure}

A sequential composition places a right-hand and a left-hand action one after the other on a target long enough for both, a walking clip, a sitting clip, or one of the two actions itself as a standing body.
The second action starts 15 frames before the first ends, the spine and head are interpolated between the two over this overlap, and outside its period each source is frozen at its first or last frame together with its object.
Since the two objects are attended to in turn, the gaze check is skipped, and the feasibility checks apply as above (rows two and four of \cref{fig:supp_composition}).

\begin{figure}[p]
    \centering
    \includegraphics[width=\linewidth,height=0.9\textheight,keepaspectratio]{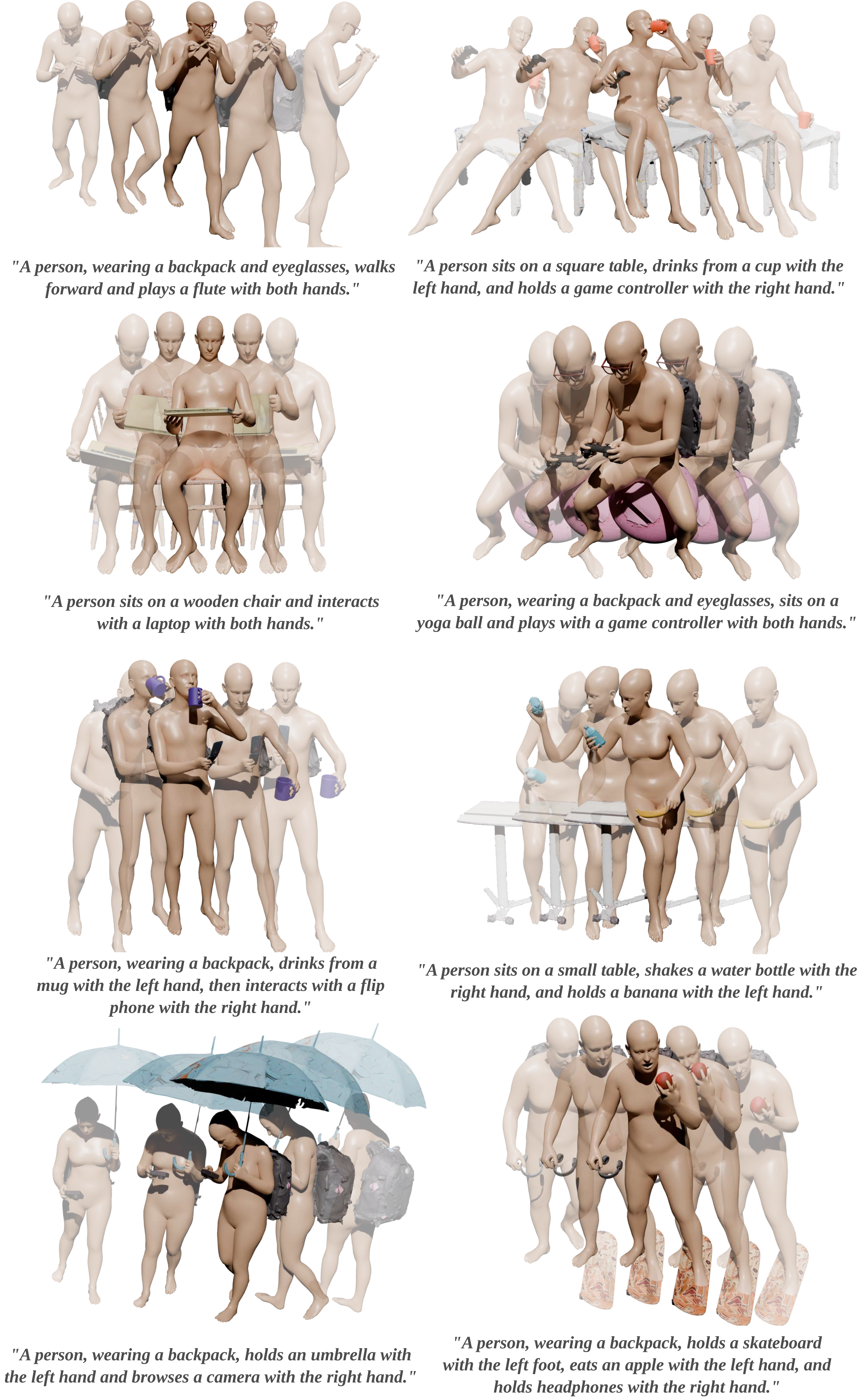}
    \caption{\textbf{Sequences of the \datasetname{}.} Composed sequences with their captions: simultaneous and sequential actions on walking, sitting, and standing bodies, with worn and articulated objects.}
    \label{fig:supp_dataset}
\end{figure}

Five steps enumerate the graph, the first feeding the one-object level: motion, walking clips composed with every compatible source, the one-hand ones serving as further targets; one object, every atomic clip and motion-step composition; two objects, every target with each of its valid sources; three objects, every two-object sequence with the sources valid for both of its rules; and sequential, the ordered pairs above.
Worn objects are added at every level.
Captions join the clause of each clip with ``and'' or, for sequential actions, ``then'', as in \emph{``A person, wearing a backpack, sits on a wooden chair and drinks from a cup with the right hand.''}
A hand-written bank paraphrases the subject, the sentence frame, the verb, and the hand phrasing, but keeps the object noun phrases verbatim.
Training draws one paraphrase per sample, and evaluation uses one fixed paraphrase per sequence (\cref{fig:supp_dataset}).

\clearpage

\section{Method Details}
\label{sec:app_method}

\subsection{Human representation}
\label{sec:app_human}
Each frame of $\h$ (\cref{eq:human_modality}) holds 256 values: the global positions $\hjoints^t \in \real^{24 \times 3}$ of the 22 body joints from the pelvis to the wrists and of the first joint of each middle finger, which fixes the wrist orientation, and the global rotations $\hpose^t \in \real^{22 \times 6}$ of the body joints in the 6D representation~\citep{zhou2019continuity}.
The PCA coefficients $\hhands^t \in \real^{2 \times 24}$ of both hands and the floor-contact scores $\hcontact^t \in \real^{4}$ of the ankles and toes complete the frame.
At decoding, the mesh is posed from the rotations and the hand coefficients with the predicted pelvis as its root, and the other positions serve as a training signal tied to the rotations by the forward-kinematics loss.

\subsection{Object representation}
\label{sec:app_object}
Each frame of $\obj$ (\cref{eq:object_modality}) holds 80 values.
The absolute pose $\objglobal^t = (\mat{R}^t_\obj, \vec{p}^t_\obj) \in \real^{9}$ is the rotation relative to the orientation of the object in the first frame of the window, in the 6D representation, and the center of mass in the canonical frame.
The hinge channel $\objhinge^t = (\cos\phi^t_\obj, \sin\phi^t_\obj)$ encodes the articulation angle relative to the first frame, constant $(1, 0)$ for rigid objects.
The relative pose with respect to a candidate joint $j \in \relj$ is
\begin{equation}
    \objrel{j}^t = \Big( (\mat{R}^t_j)^{\top} \mat{R}^t_\obj,\; (\mat{R}^t_j)^{\top} (\vec{p}^t_\obj - \vec{p}^t_j) / r_\obj \Big) \in \real^{9}, \qquad j \in \relj,
    \label{eq:relative_stream}
\end{equation}
where $\mat{R}^t_j$ and $\vec{p}^t_j$ come from $\hpose^t$ and $\hjoints^t$ and $r_\obj$ is the bounding radius of the object plus 0.5\,m.
The normalized translation then lies in $[-1, 1]$ whenever the joint is near the bounding sphere, the only frames in which the stream is supervised.
The contact scores $\objcontact^t \in \real^{15}$ of the ten fingers, the feet, the pelvis, the spine, and the head are soft indicators $\sigma\big({-}100\,(d - \delta)\big)$ of the distance $d$ in meters to the object surface.
Positions and hand coefficients are min-max normalized, and contact scores are scaled to $[-1, 1]$.
An articulated object occupies one slot with the pose of its base part and the hinge angle, and its geometry is the assembled mesh at the hinge angle of the first frame.

\subsection{Geometry encoder}
\label{sec:app_geometry}
The BPS representation~\citep{prokudin2019efficient} describes an object by the offsets from 1024 fixed basis points, sampled in a ball around its center of mass, to the nearest points of its surface, for the mesh oriented as in the first frame of the window.
Our encoder aggregates these offsets over a fixed partition of the ball into 8 horizontal slabs of 16 cells each, split by azimuth and by distance to the vertical axis.
Per-point features from a shared MLP are pooled within every cell and then within every slab, and the slab features are concatenated from bottom to top and projected to the geometry embedding, which preserves the vertical layout of the object.

\subsection{Architecture}
\label{sec:app_architecture}
The denoiser consists of eight pre-LN transformer blocks with a model width of 512, eight attention heads, a feed-forward width of 2048, and a dropout of 0.1.
Every human frame and every chunk of four object frames is a token, so a window with five objects has 270 motion tokens instead of 720.
A one-dimensional convolution of kernel size three decodes the object tokens to per-frame parameters.
The temporal order is encoded with rotary position embeddings~\citep{su2024roformer}: a human frame and an object chunk at the same time receive the same position, whereas the step, sentence, and text tokens receive position zero.
A learned type embedding distinguishes human, object, sentence, and text tokens.
The denoiser has 27.3M parameters, 25.2M of them in the transformer blocks, the part-assignment head 2.6M, the geometry encoder 0.15M, and the CLIP projection 0.26M, 30.4M in total.
The frozen CLIP ViT-B/32 text encoder adds 63.4M parameters, and our re-implementations of MDM and PriorMDM have 20.7M and 39.6M.

\subsection{Part-assignment head}
\label{sec:app_assignment}
Per slot, the head receives the object conditioning, the initial pose of the person, and the sentence embedding, then two blocks of self-attention across slots and cross-attention to the text tokens produce one binary output per candidate joint.
A joint is labeled as driving an object when it touches the object for at least ten frames and the object travels at least half of the distance that the joint travels meanwhile, and the denoiser receives these labels during training.
At decoding, the pose source is selected once per window: the driving joint with the highest average predicted score, a wrist scored by the maximum over its fingers.
If no joint is predicted as driving or the highest score is below 0.5, the absolute pose is used.
The baselines, which have no part-assignment head, select among all six candidate joints by the same contact scores, so an object follows a joint only while its predicted contact stays high.

\subsection{Loss terms}
\label{sec:app_losses}
The terms $\loss{k}$ of \cref{eq:objective} are $\ell_1$ errors of the human parameters and of their frame-to-frame differences, of the absolute object pose, of the relative poses in their supervised frames, and of the hinge channel, and a cross-stream consistency term.
Further terms are squared errors of the object and foot-floor contact scores and a foot-skating penalty on the predicted ankles and toes in frames where the ground-truth foot is in contact with the floor.
A forward-kinematics term compares the joint positions of the predicted rotations with the ground-truth positions, and the binary cross-entropy of the part-assignment head completes the objective.
The consistency term poses 100 surface points of the object with each of the seven predicted poses, points on a moving part with the predicted hinge angle, and penalizes the $\ell_1$ distance of the points and of their frame-to-frame differences to the ground truth.
All weights $\lambda_k$ are 1, except 2 for the human-parameter term.

\subsection{Training details}
\label{sec:app_training}
We train for 80k steps with batches of 256 windows using Adam~\citep{kingma2015adam} with a learning rate of $5{\cdot}10^{-4}$ and a cosine schedule with warm-up, and sample from an exponential moving average of the weights.
Training takes eight hours on one RTX 5090.
The text prompt is dropped with probability 0.1, replacing the sentence token by a learned null embedding and masking the text tokens.
Each worn object is included with probability 0.5, an object name by a null embedding with probability 0.3, and the part-assignment input is removed with probability 0.1.

\input{tables/09_runtime}

\subsection{Runtime}
\label{sec:app_runtime}
\Cref{tab:runtime} reports generation times on one NVIDIA RTX 5090 with the sampler of \cref{sec:method}, in fp16 with a compiled denoiser, where the conditional and unconditional passes of classifier-free guidance run as one doubled batch.
Single sequences are timed at batch size one as the median of 20 runs after three warm-up runs.
The median test sequence has 119 frames and fits one window.
The denoising steps dominate without the guidance and its Adam iterations with it.
The guidance therefore costs 7.2 times the generation time of a single sequence, but only 1.55 times in the batched evaluation of the test set, since its small kernels underuse the GPU on one sequence.

\subsection{Long sequences}
\label{sec:app_long}
Consecutive windows of 120 frames overlap by 60 frames.
The human pose and the object states predicted for the first frame of a window define its canonical frame and the reference orientation and hinge angle of every object.
In the overlap, the two predictions are blended with linear weights, rotations projected back onto $SO(3)$ and hinge angles averaged as $(\cos, \sin)$ pairs.

\subsection{Contact-consistent guidance}
\label{sec:app_guidance}
The two guidance losses of \cref{sec:method} invert the contact labels.
The score of finger $f$ and object $\obj_n$ in $\objcontact^t$, which is the soft indicator defined above, and the floor-contact score of foot joint $j$ in $\hcontact^t$ are
\begin{equation}
\begin{aligned}
    c^{t}_{\obj_n, f} &= \sigma\big({-}100\,(d^{t}_{\obj_n, f} - \delta)\big), \\
    c^{t}_{\h, j} &= \sigma\big({-}100\,(y^{t}_j - \delta_y)\big)\, \sigma\big({-}800\,(v^{t}_j - \delta_v)\big),
\end{aligned}
    \label{eq:contact_labels}
\end{equation}
where $d^{t}_{\obj_n, f}$ is the distance between the finger and the object surface, $y^{t}_j$ the height of the foot joint above the floor, and $v^{t}_j = \lVert \vec{p}^{t+1}_j - \vec{p}^{t}_j \rVert$ its displacement to the next frame, all in meters.
The slopes of 100 and 800 per meter set the width of the transition around corresponding $\delta$ thresholds.
Both labels are monotone, so a score $c$ implies the bounds
\begin{equation}
    d \le d^*(c) = \delta - \tfrac{1}{100}\, \mathrm{logit}(c),
    \qquad
    v \le v^*(c) = \delta_v - \tfrac{1}{800}\, \mathrm{logit}(c),
    \label{eq:guidance_bounds}
\end{equation}
which hold exactly on the training data.
For every score that predicts a contact, $c > 0.5$, the losses penalize the violation of its bound,
\begin{equation}
    \loss{contact} = \sum_{n, f, t} c^{t}_{\obj_n, f}\, \big[ d^{t}_{\obj_n, f} - d^*\big(c^{t}_{\obj_n, f}\big) \big]_+ ,
    \qquad
    \loss{slide} = \sum_{j, t} c^{t}_{\h, j}\, \big[ v^{t}_j - v^*\big(c^{t}_{\h, j}\big) \big]_+ ,
    \label{eq:guidance}
\end{equation}
where $[\cdot]_+ = \max(\cdot, 0)$, and we minimize $\loss{contact} + \loss{slide}$.
The guidance runs at the last ten of the 100 sampling steps, with five Adam iterations at a learning rate of $5{\cdot}10^{-4}$ on the predicted clean sample at each of them and 100 at the final step, whose prediction is the output.
Within every iteration, the edit is blurred along time with a Gaussian kernel of one frame, to reduce jitter.

%% file: tables/02_dataset_metrics_parents.tex
\begin{table}[t]
    \centering
    \caption{\textbf{Quality of composed sequences.} Physical plausibility of the captured parent sequences of every source (Original) and of the composed sequences built from them (\methodname{}), measured with the metrics of HUMOTO~\citep{lu2025humoto}. The difference to the parents is given in parentheses and colored by its relative size: green for an improvement, amber for an increase below 50\,\%, red for an increase of 50\,\% or more, and gray for a negligible change. Metrics that do not apply to a source are marked with \na.}
    \label{tab:dataset-metrics-parents}
    \footnotesize
    \setlength{\tabcolsep}{5pt}
    \renewcommand{\arraystretch}{1.2}
    \begin{adjustbox}{max width=\linewidth}
\begin{tabular}{l cc cc cc cc cc}
    \toprule
    \multirow{3}{*}{\makecell[l]{\textbf{Dataset}\\[2pt]{\footnotesize\itshape (parent sequences only)}}}
      & \multicolumn{4}{c}{\textbf{Human Motion}}
      & \multicolumn{2}{c}{\textbf{Object Motion}}
      & \multicolumn{4}{c}{\textbf{Human-Object Penetration}} \\
    \cmidrule(lr){2-5} \cmidrule(lr){6-7} \cmidrule(lr){8-11}
      & \multicolumn{2}{c}{\textbf{Foot sliding (cm) $\downarrow$}}
      & \multicolumn{2}{c}{\textbf{Jerk (m/s$^{3}$) $\downarrow$}}
      & \multicolumn{2}{c}{\textbf{Jerk (m/s$^{3}$) $\downarrow$}}
      & \multicolumn{2}{c}{\textbf{Body (cm) $\downarrow$}}
      & \multicolumn{2}{c}{\textbf{Hands (cm) $\downarrow$}} \\
    \cmidrule(lr){2-3} \cmidrule(lr){4-5} \cmidrule(lr){6-7} \cmidrule(lr){8-9} \cmidrule(lr){10-11}
      & \textbf{Original} & \textbf{COSMI}
      & \textbf{Original} & \textbf{COSMI}
      & \textbf{Original} & \textbf{COSMI}
      & \textbf{Original} & \textbf{COSMI}
      & \textbf{Original} & \textbf{COSMI} \\
    \midrule
    GRAB~\citep{taheri2020grab}        & 0.03 & \cw{0.03}{\dgray{0.00}} & 0.33 & \cw{0.44}{\dmild{0.11}} & 1.67 & \cw{1.37}{\dgood{0.30}} & 0.08 & \cw{0.05}{\dgood{0.03}} & 0.54 & \cw{0.53}{\dgood{0.01}} \\
    BEHAVE~\citep{bhatnagar2022behave} & 0.13 & \cw{0.11}{\dgood{0.02}} & 1.15 & \cw{1.11}{\dgood{0.04}} & 1.71 & \cw{2.03}{\dmild{0.32}} & 1.64 & \cw{1.93}{\dmild{0.29}} & \na & \na \\
    ARCTIC~\citep{fan2023arctic}       & 0.01 & \cw{0.01}{\dgray{0.00}} & 0.31 & \cw{0.45}{\dmild{0.14}} & 1.16 & \cw{1.23}{\dmild{0.07}} & 0.00 & \cw{0.01}{\dgray{0.01}} & 0.79 & \cw{1.03}{\dmild{0.24}} \\
    InterCap~\citep{huang2024intercap} & 0.08 & \cw{0.09}{\dmild{0.01}} & 1.96 & \cw{2.34}{\dmild{0.38}} & 2.35 & \cw{2.58}{\dmild{0.23}} & 0.17 & \cw{0.40}{\dsev{0.23}} & 0.52 & \cw{0.23}{\dgood{0.29}} \\
    OMOMO~\citep{li2023omomo}          & 0.29 & \cw{0.29}{\dgray{0.00}} & 1.25 & \cw{1.32}{\dmild{0.07}} & 1.44 & \cw{1.35}{\dgood{0.09}} & 0.01 & \cw{0.01}{\dgray{0.00}} & \na & \na \\
    AMASS~\citep{mahmood2019amass}     & 0.17 & \cw{0.18}{\dmild{0.01}} & 2.56 & \cw{3.42}{\dmild{0.86}} & \na & \cw{5.30}{} & \na & \na & \na & \na \\
    \bottomrule
  \end{tabular}
    \end{adjustbox}
\end{table}

%% file: tables/05_ablations.tex
\begin{table}[t]
    \centering
    \caption{\textbf{Ablations.} Each row changes one component of the full model. The row w/o guidance samples without the contact-consistent guidance of \cref{sec:method}, and all rows below it are sampled without guidance as well. Of these, w/o hier.\ enc.\ replaces the hierarchical geometry encoder by an MLP, $\obj$ chunk $n$ uses $n$ frames per object token instead of 4, and aux.\ loss denotes the interaction and plausibility terms. The auxiliary losses matter most, part assignment improves the alignment with the text and the contacts, and text tokens and object names improve the alignment with the text. Format as in \cref{tab:comparison_full}.}
    \label{tab:ablations}
    \footnotesize
    \setlength{\tabcolsep}{2pt}
    \begin{adjustbox}{max width=\linewidth}
\begin{tabular}{@{}lcccc@{\hspace{0.8em}}c@{\hspace{0.8em}}c@{\hspace{0.8em}}ccc@{}}
    \toprule
    \multirow{2}{*}{\bf Method}
        & \multicolumn{4}{c@{\hspace{0.8em}}}{\bf Text-to-motion}
        & \multicolumn{1}{c@{\hspace{0.8em}}}{\bf $\h$}
        & \multicolumn{1}{c@{\hspace{0.8em}}}{\bf $\obj$}
        & \multicolumn{3}{c@{}}{\bf Contact} \\
    \cmidrule(lr){2-5} \cmidrule(lr){6-6} \cmidrule(lr){7-7} \cmidrule(lr){8-10}
        & \overmean{FID$\downarrow$}{0.000}\noerr{.000}
        & \overmean{Top-1$\uparrow$}{0.000}\noerr{.000}
        & \overmean{Top-3$\uparrow$}{0.000}\noerr{.000}
        & \overmean{MM-Dist$\downarrow$}{0.000}\noerr{.000}
        & Foot sl.$\downarrow$
        & Jerk$\downarrow$
        & T-Acc$\uparrow$
        & B-Acc$\uparrow$
        & Pen. (\%)$\downarrow$ \\
    \midrule
    Data
        & \nomean{0.049}\noerr{.000} & 0.740\noerr{.000}        & 0.919\noerr{.000}       & 0.740\noerr{.000}
        & 1.879
        & 0.366
        & --            & --             & 34.60 \\
    \midrule
    \methodname{} (ours)
        & 0.049\err{.000}            & \second{0.696}\err{.003} & \first{0.900}\err{.001} & \second{0.827}\err{.002}
        & \first{1.395}
        & 0.409
        & \first{0.980} & 0.968          & 33.96 \\
    w/o guidance
        & 0.050\err{.000}            & \first{0.697}\err{.003}  & \first{0.900}\err{.001} & \first{0.825}\err{.002}
        & 1.744
        & 0.397
        & 0.979         & 0.970          & \second{33.55} \\
    \midrule
    w/o hier.\ enc.
        & \second{0.041}\err{.000}   & 0.695\err{.001}          & 0.893\err{.001}         & 0.845\err{.000}
        & 1.757
        & 0.364
        & 0.975         & \second{0.971} & 33.62 \\
    w/o names
        & 0.057\err{.000}            & 0.687\err{.001}          & 0.889\err{.001}         & 0.877\err{.001}
        & 1.768
        & 0.383
        & 0.974         & 0.969          & 34.84 \\
    $\obj$ chunk 1
        & 0.052\err{.000}            & 0.686\err{.000}          & 0.883\err{.001}         & 0.886\err{.001}
        & 1.658
        & \first{0.303}
        & 0.957         & \first{0.974}  & \first{33.12} \\
    $\obj$ chunk 8
        & \first{0.039}\err{.001}    & 0.677\err{.001}          & 0.881\err{.000}         & 0.903\err{.002}
        & 1.714
        & 0.393
        & 0.979         & 0.970          & 33.66 \\
    w/o $\textok$
        & 0.043\err{.000}            & 0.675\err{.001}          & 0.888\err{.000}         & 0.907\err{.003}
        & \second{1.637}
        & \second{0.334}
        & \first{0.980} & 0.968          & 33.91 \\
    w/o part assign.
        & 0.069\err{.000}            & 0.644\err{.001}          & 0.859\err{.002}         & 0.982\err{.003}
        & 1.727
        & 0.405
        & 0.945         & 0.944          & 34.47 \\
    w/o aux.\ loss
        & 0.256\err{.002}            & 0.550\err{.002}          & 0.783\err{.001}         & 1.435\err{.003}
        & 3.633
        & 0.449
        & 0.810         & 0.910          & 36.25 \\
    \bottomrule
\end{tabular}
    \end{adjustbox}
\end{table}

%% file: tables/08_himo_dataset.tex
\begin{table}[t]
    \centering
    \caption{\textbf{Comparison on the HIMO dataset.} Values of the baselines as reported by HIMO~\citep{lv2024himo}, with the 95\,\% confidence intervals that HIMO reports, and the Data row is HIMO's Real row. \methodname{} is trained on the HIMO dataset and scored with the evaluator of HIMO over 20 replications, sampled without guidance, as the closest setting to that of HIMO. Format as in \cref{tab:comparison_full}.}
    \label{tab:himo_dataset}
    \footnotesize
    \setlength{\tabcolsep}{2pt}
    \begin{adjustbox}{max width=\linewidth}
\begin{tabular}{@{}clccccc@{}}
    \toprule
    & \multirow{2}{*}{\bf Method}
        & \multicolumn{5}{c@{}}{\bf Text-to-motion} \\
    \cmidrule(lr){3-7}
    &
        & \overmean{FID$\downarrow$}{0.000}\noerr{.000}
        & \overmean{Top-3$\uparrow$}{0.000}\noerr{.000}
        & \overmean{MM-Dist$\downarrow$}{0.000}\noerr{.000}
        & \overmean{Div.$\rightarrow$}{0.00}\noerr{.00}
        & \overmean{MModal.$\uparrow$}{0.000}\noerr{.000} \\
    \midrule
    \multirow{5}{*}{\rotatebox[origin=c]{90}{2 obj.}}
    & Data
        & 0.018\err{.006}          & 0.799\err{.008}          & 3.566\err{.011}          & 11.40\err{.26}          & \nomean{1.288}\noerr{.000} \\
    \cmidrule{2-7}
    & MDM
        & 6.846\err{.332}          & 0.605\err{.010}          & 8.019\err{.050}          & \first{11.39}\err{.23}  & 1.288\err{.211} \\
    & PriorMDM
        & 7.852\err{.252}          & 0.589\err{.003}          & 7.251\err{.006}          & 12.58\err{.15}          & \second{1.591}\err{.145} \\
    & HIMO
        & \first{1.481}\err{.043}  & \second{0.637}\err{.003} & \first{3.649}\err{.010}  & \second{11.66}\err{.20} & \first{1.786}\err{.057} \\
    \cmidrule{2-7}
    & \methodname{} (ours)
        & \second{4.896}\err{.038} & \first{0.730}\err{.005}  & \second{4.436}\err{.010} & 10.76\err{.12}          & 0.691\err{.015} \\
    \midrule\midrule
    \multirow{5}{*}{\rotatebox[origin=c]{90}{3 obj.}}
    & Data
        & 0.181\err{.044}          & 0.699\err{.005}          & 3.770\err{.028}          & 9.77\err{.22}           & \nomean{0.789}\noerr{.000} \\
    \cmidrule{2-7}
    & MDM
        & \second{4.571}\err{.110} & 0.502\err{.014}          & 6.314\err{.027}          & 8.90\err{.29}           & \second{0.789}\err{.572} \\
    & PriorMDM
        & 4.821\err{.203}          & 0.514\err{.026}          & 5.890\err{.023}          & \second{9.34}\err{.02}  & \first{0.812}\err{.328} \\
    & HIMO
        & 4.771\err{.164}          & \second{0.535}\err{.018} & \second{5.087}\err{.041} & 8.95\err{.14}           & 0.756\err{.055} \\
    \cmidrule{2-7}
    & \methodname{} (ours)
        & \first{2.815}\err{.043}  & \first{0.582}\err{.011}  & \first{4.366}\err{.022}  & \first{9.63}\err{.14}   & 0.741\err{.028} \\
    \bottomrule
\end{tabular}
    \end{adjustbox}
\end{table}

%% file: tables/10_rules.tex
\begin{table}[!t]
    \centering
    \caption{\textbf{Interaction rules.} Contact groups, pose constraints, local reference joints, and source datasets of the thirteen rules, with the number of atomic clips each yields before and after mirroring. The walking rule covers standing as well and selects whole AMASS sequences by their BABEL labels. T: target, S: source, A: augmentation.}
    \label{tab:rules}
    \footnotesize
    \setlength{\tabcolsep}{4pt}
    \renewcommand{\arraystretch}{1.15}
    \begin{adjustbox}{max width=\linewidth}
    \begin{tabular}{@{}lllllllr@{}}
        \toprule
        \textbf{Rule} & \textbf{Interaction} & \textbf{Contact} & \textbf{No contact} & \textbf{Pose constraint} & \textbf{Reference $\bodyj$} & \textbf{Datasets} & \textbf{Clips} \\
        \midrule
        T:walk       & Walking or standing                    & \na         & \na         & \na                                & \na          & AMASS            & 41 / 41 \\
        T:sit        & Sitting on a support object            & lower back  & lower front & upright torso                      & pelvis       & BEHAVE, InterCap & 78 / 78 \\
        T:rfoot      & Right-foot manipulation                & right foot  & left foot   & upright torso                      & right foot   & InterCap         & 19 / 24 \\
        T:lfoot      & Left-foot manipulation                 & left foot   & right foot  & upright torso                      & left foot    & InterCap         & 5 / 24 \\
        T:rhand      & Right-hand manipulation, large object  & right hand  & left hand   & upright torso                      & right wrist  & InterCap, OMOMO  & 492 / 866 \\
        T:lhand      & Left-hand manipulation, large object   & left hand   & right hand  & upright torso                      & left wrist   & InterCap, OMOMO  & 374 / 866 \\
        \midrule
        S:rhand      & Right-hand manipulation, small object  & right hand  & left hand   & upright torso                      & right wrist  & GRAB, ARCTIC     & 68 / 71 \\
        S:lhand      & Left-hand manipulation, small object   & left hand   & right hand  & upright torso                      & left wrist   & GRAB, ARCTIC     & 3 / 71 \\
        S:hands      & Two-handed manipulation                & both hands  & \na         & upright torso                      & both wrists  & GRAB, ARCTIC     & 187 / 187 \\
        S:rhold      & Passive right-hand hold                & right hand  & left hand   & upright torso, elbow at $90^\circ$ & right wrist  & GRAB             & 153 / 211 \\
        S:lhold      & Passive left-hand hold                 & left hand   & right hand  & upright torso, elbow at $90^\circ$ & left wrist   & GRAB             & 58 / 211 \\
        \midrule
        A:backpack   & Backpack worn on the back              & upper back  & hands       & upright torso, arms down           & spine        & BEHAVE           & 11 / 11 \\
        A:eyeglasses & Eyeglasses worn on the face            & face        & hands       & \na                                & both eyes    & GRAB             & 2 / 2 \\
        \midrule
        \multicolumn{7}{@{}l}{Total} & 1,491 / 2,663 \\
        \bottomrule
    \end{tabular}
    \end{adjustbox}
\end{table}

%% file: tables/09_runtime.tex
\begin{wraptable}{r}{0.5\linewidth}
    \vspace{-12pt}
    \centering
    \caption{\textbf{Runtime.} Generation time on one RTX 5090 without and with the contact-consistent guidance (G), at batch size one unless stated.}
    \label{tab:runtime}
    \footnotesize
    \setlength{\tabcolsep}{4pt}
    \renewcommand{\arraystretch}{1.15}
    \begin{tabular}{@{}l r r@{}}
        \toprule
        & \textbf{w/o G} & \textbf{w/ G} \\
        \midrule
        \textit{Per window (ms)}      &       &       \\
        \quad total                   & 157   & 1,140 \\
        \quad denoising, 100 steps    & 138   & 1,122 \\
        \quad decoding to meshes      & 5     & 5     \\
        \textit{Per sequence (s)}     &       &       \\
        \quad 119 frames, 1 window    & 0.15  & 1.08  \\
        \quad 299 frames, 4 windows   & 0.63  & 4.56  \\
        Batched (s per seq., batch 128) & 0.088 & 0.136 \\
        Peak GPU memory (GiB)         & 0.78  & 0.97  \\
        \bottomrule
    \end{tabular}
    \vspace{-8pt}
\end{wraptable}